\documentclass[lettersize,journal]{IEEEtran}
\usepackage{amsmath,amsfonts}
\usepackage{algorithmic}
\usepackage{algorithm}
\usepackage{array}
\usepackage[caption=false,font=normalsize,labelfont=sf,textfont=sf]{subfig}
\usepackage{textcomp}
\usepackage{stfloats}
\usepackage{url}
\usepackage{verbatim}
\usepackage{graphicx}
\usepackage{cite}
\usepackage[hidelinks]{hyperref}
\usepackage{pdfpages}
\usepackage{cite}
\usepackage{url}
\usepackage[flushleft]{threeparttable}
\usepackage{bm}
\usepackage{tabularx}
\usepackage{graphicx}
\usepackage{subfig}
\usepackage{float}
\usepackage{footnote}
\makesavenoteenv{tabular}
\makesavenoteenv{table}
\usepackage{amsmath}
\usepackage{booktabs}
\usepackage{caption}
\usepackage{amssymb}
\usepackage[capitalise]{cleveref}
\usepackage{algorithm}
\usepackage{algorithmic}
\usepackage{multirow}
\usepackage{multicol}
\usepackage{tablefootnote}
\usepackage{graphicx}
\usepackage{subcaption} 

\usepackage{tikz}
\usetikzlibrary{decorations.markings}
\usepackage{pgfplots}
\pgfplotsset{compat=newest}

\tikzset{
    car_top/.pic={
        \filldraw [black, fill=#1, very thin] plot[smooth, black, tension=.7, very thin] 
            coordinates {(2.3248,0) (2.2967,0.4058) (2.0941,0.7925)(1.7847,0.9958) (1.1874,1.0261) (0.9497,1.0108)}
            .. controls (0.9124,1.143) and (0.8648,1.2352) .. (0.7463,1.1888) .. controls (0.7628,1.0703) and (0.7832,1.0089) .. (0.8202,0.952) .. controls (0.6029,0.9902) and (0.1582,1.0079) .. (-0.2536,0.9816) .. controls (-0.3325,1.0605) and (-0.4289,1.0459) .. (-0.5136,0.9787) .. controls (-0.6947,0.9904) and (-0.8232,0.9758) .. (-0.94,0.9524)
            plot[smooth, tension=.7,very thin] coordinates {(-0.94,0.9524)(-1.1951,1.0129) (-1.8081,0.9951) (-2.1686,0.8328) (-2.3894,0.4889) (-2.4286,0) };

        \fill [fill=#1,very thin](-2.4286,0)--(-0.94,0.9524)--(2.3248,0)--(-0.94,-0.9524);

        \filldraw  [black, fill=#1, very thin] plot[smooth, black, tension=.7,very thin] 
            coordinates {(2.3248,0) (2.2967,-0.4058) (2.0941,-0.7925)(1.7847,-0.9958) (1.1874,-1.0261) (0.9497,-1.0108)}
            .. controls (0.9124,-1.143) and (0.8648,-1.2352) .. (0.7463,-1.1888) .. controls (0.7628,-1.0703) and (0.7832,-1.0089) .. (0.8202,-0.952) .. controls (0.6029,-0.9902) and (0.1582,-1.0079) .. (-0.2536,-0.9816) .. controls (-0.3325,-1.0605) and (-0.4289,-1.0459) .. (-0.5136,-0.9787) .. controls (-0.6947,-0.9904) and (-0.8232,-0.9758) .. (-0.94,-0.9524)
            plot[smooth, tension=.7,very thin] coordinates {(-0.94,-0.9524)(-1.1951,-1.0129) (-1.8081,-0.9951) (-2.1686,-0.8328) (-2.3894,-0.4889) (-2.4286,0) };

        \draw [smooth cycle, black, fill=darkgray,very thin](0.52,0.75) .. controls (0.7,0.78) and (1,0.82) .. (1.2,0.85) .. controls (1.7,0.5) and (1.7,-0.5) .. (1.2,-0.85) .. controls (1,-0.82) and (0.7,-0.78) .. (0.52,-0.75) .. controls (0.7,-0.25) and (0.7,0.25) .. (0.52,0.75);
        \draw [smooth cycle, black, fill=darkgray,very thin](-1.06,0.67) .. controls (-1.33,0.68) and (-1.57,0.67) .. (-1.8,0.68) .. controls (-2.1,0.27) and (-2.1,-0.27) .. (-1.8,-0.68) .. controls (-1.57,-0.67) and (-1.33,-0.68) .. (-1.06,-0.67) .. controls (-1.16,-0.3) and (-1.16,0.3) .. (-1.06,0.67);
        \draw [smooth cycle, black, fill=darkgray,very thin](0.97,0.87) .. controls (0.32,1) and (-0.84,0.93) .. (-1.47,0.83) .. controls (-1.23,0.8) and (-1.0974,0.7863) .. (-0.88,0.76) .. controls (-0.56,0.75) and (0.02,0.76) .. (0.4,0.78) .. controls (0.63,0.82) and (0.7,0.83) .. (0.97,0.87);
        \draw [smooth cycle, black, fill=darkgray,very thin](0.97,-0.87) .. controls (0.32,-1) and (-0.84,-0.93) .. (-1.47,-0.83) .. controls (-1.23,-0.8) and (-1.0974,-0.7863) .. (-0.88,-0.76) .. controls (-0.56,-0.75) and (0.02,-0.76) .. (0.4,-0.78) .. controls (0.63,-0.82) and (0.7,-0.83) .. (0.97,-0.87);

        \draw [rotate=51, black, fill=red, very thin] (-0.75,2.07) ellipse (0.18 and 0.06);
        \draw [rotate=-51, black, fill=red, very thin] (-0.75,-2.07) ellipse (0.18 and 0.06);
        \draw [rotate=-51, black, fill=white, very thin] (0.6696,2.0406) ellipse (0.19 and 0.05);
        \draw [rotate=51, black, fill=white, very thin] (0.6696,-2.0406) ellipse (0.19 and 0.05);
    }
}

\begin{document}

\title{Mode Collapse Happens: Evaluating Critical Interactions in Joint Trajectory Prediction Models}


\author{
    Maarten Hugenholtz,
    Anna Meszaros,
    Jens Kober,
    Zlatan Ajanovic
    \thanks{M. Hugenholtz, A. Meszaros, and J. Kober are with the Cognitive Robotics Department, Delft University of Technology, 2628 CD Delft, The Netherlands (e-mail: M.D.Hugenholtz@student.tudelft.nl; A.Meszaros@tudelft.nl; J.Kober@tudelft.nl).}
    \thanks{Z. Ajanovic is with the Computer Science Department, RWTH Aachen University, 52062 Aachen, Germany (e-mail: Zlatan.Ajanovic@ml.rwth-aachen.de).}
}


\markboth{Journal of \LaTeX\ Class Files,~Vol.~x, No.~x, April~2025}%
{Shell \MakeLowercase{\textit{et al.}}: A Sample Article Using IEEEtran.cls for IEEE Journals}

\IEEEpubid{0000--0000/00\$00.00~\copyright~2024 IEEE}

\maketitle

\begin{abstract}
Autonomous Vehicle decisions rely on multimodal prediction models that account for multiple route options and the inherent uncertainty in human behavior. However, models can suffer from mode collapse, where only the most likely mode is predicted, posing significant safety risks. While existing methods employ various strategies to generate diverse predictions, they often overlook the diversity in interaction modes among agents. Additionally, traditional metrics for evaluating prediction models are dataset-dependent and do not evaluate inter-agent interactions quantitatively. To our knowledge, none of the existing metrics explicitly evaluates mode collapse.
In this paper, we propose a novel evaluation framework that assesses mode collapse in joint trajectory predictions, focusing on safety-critical interactions. We introduce metrics for mode collapse, mode correctness, and coverage, emphasizing the sequential dimension of predictions. By testing four multi-agent trajectory prediction models, we demonstrate that mode collapse indeed happens. When looking at the sequential dimension, although prediction accuracy improves closer to interaction events, there are still cases where the models are unable to predict the correct interaction mode, even just before the interaction mode becomes inevitable.
We hope that our framework can help researchers gain new insights and advance the development of more consistent and accurate prediction models, thus enhancing the safety of autonomous driving systems.

\end{abstract}



\section{INTRODUCTION}
\IEEEPARstart{A}{utonomous} vehicles (AVs) have the potential to revolutionize personal transportation, motivated by improved driving comfort, energy efficiency and road safety \cite{othman_exploring_2022}. 
Part of the autonomous driving challenge involves the planning of safe, comfortable and efficient driving trajectories. 
To achieve this, modular planning systems rely on a prediction module that predicts the motion of surrounding vehicles \cite{hagedorn_rethinking_2023}. 
As human behavior is naturally uncertain and multimodal, it is unrealistic to predict a single trajectory for each agent, without knowing the agent’s intent. Therefore, multimodal trajectory prediction (MTP) was introduced by Gupta et al. \cite{gupta_social_2018}, where multiple trajectories are predicted for each agent, to cover different possible modes. 

However, a common problem in multimodal trajectory prediction models is their susceptibility to mode collapse. 
This machine learning phenomenon occurs when the model fails to learn the true distribution of modes and only outputs the most likely mode, or two distant modes collapse into a single, infeasible mode \cite{amirian_social_2019}. 
In a safety-critical application like autonomous driving, it is crucial that such failures are avoided, as incomplete or inaccurate predictions, that are used in a downstream planner, could result in collisions.
Several works try to address the mode collapse issue either by using goal-conditioned prediction and a diverse set of goals \cite{zhao_tnt_2021, gu_densetnt_2021, chen_categorical_2023}, by using training objectives that allow for diverse predictions \cite{chen_scept_2022, yuan_dlow_2020} or by promoting distributions with high entropy \cite{deo_trajectory_2021}. 
Generally, these works consider mode collapse on the environmental level by generating diverse predictions that cover various route options, 
but little attention has been given to ensuring diversity in the interaction modes among agent trajectories (e.g. in the spatiotemporal domain \cite{ajanovic_search-based_2018}). 
Furthermore, there are currently no metrics to evaluate mode collapse explicitly. 

\begin{figure}
    \centering
    \subfloat[{\scriptsize Ground truth}]{\resizebox{0.31\linewidth}{!}{    
            \begin{tikzpicture}
    \fill[gray!30] (-3,-3) rectangle (3,3);
    \fill[white] (-1,-3) rectangle (1,3);
    \fill[white] (-3,-1) rectangle (3,1);

    \pic[scale=0.15] at (-2.5,-0.5) {car_top={red}};
    \pic[scale=0.15, rotate=180] at (-1,0.5) {car_top={darkgray}};
    \pic[scale=0.15, rotate=-270] at (0.5,-2.5) {car_top={green}};

    \draw[->, thick, blue, decoration={markings, mark=at position 0.5 with {\arrow[scale=1.5]{>}}}]
        (0.5,-2.5) .. controls (0.5,-2)  .. (0.5, -1);

    \draw[->, thick, blue, decoration={markings, mark=at position 0.5 with {\arrow[scale=1.5]{>}}}]
        (-2.5,-0.5) .. controls (0,-0.5)  .. (2.8,-0.5);

    \draw[->, thick, blue, decoration={markings, mark=at position 0.5 with {\arrow[scale=1.5]{>}}}]
        (-1,0.5) .. controls(-2,0.5)  .. (-2.8,0.5);




    \node[above, black, font=\small] at (-2.1,-2) {GT};
    \draw[->, thick, blue, decoration={markings, mark=at position 0.5 with {\arrow[scale=1.5]{>}}}]
    (-2.8,-2) .. controls(-2,-2)  .. (-1.3,-2);

    \node[above, black, font=\small] at (-2.1,-2.7) {Prediction};
    \draw[->, thick, dotted, purple, decoration={markings, mark=at position 0.5 with {\arrow[scale=1.5]{>}}}]
    (-2.8,-2.7) .. controls(-2,-2.7)  .. (-1.3,-2.7);
    
\end{tikzpicture}  
        }
    }
    \subfloat[{\scriptsize Correct mode}]{ \resizebox{0.31\linewidth}{!}{%
            \begin{tikzpicture}
    \fill[gray!30] (-3,-3) rectangle (3,3);
    \fill[white] (-1,-3) rectangle (1,3);
    \fill[white] (-3,-1) rectangle (3,1);

    \pic[scale=0.15] at (-2.5,-0.5) {car_top={red}};
    \pic[scale=0.15, rotate=180] at (-1,0.5) {car_top={darkgray}};
    \pic[scale=0.15, rotate=-270] at (0.5,-2.5) {car_top={green}};




    \draw[->, thick, dotted, purple, decoration={markings, mark=at position 0.5 with {\arrow[scale=1.5]{>}}}]
        (0.5,-2.5) .. controls (0.45,-2)  .. (0.4, -1.5);

    \draw[->, thick, dotted, purple, decoration={markings, mark=at position 0.5 with {\arrow[scale=1.5]{>}}}]
        (-2.5,-0.5) .. controls (0,-0.45)  .. (1,-0.35);

    \draw[->, thick, dotted, purple, decoration={markings, mark=at position 0.5 with {\arrow[scale=1.5]{>}}}]
        (-1,0.5) .. controls(-1.5,0.55)  .. (-2.3,0.6);

    \node[above, black, font=\small] at (-2.1,-2) {GT};
    \draw[->, thick, blue, decoration={markings, mark=at position 0.5 with {\arrow[scale=1.5]{>}}}]
    (-2.8,-2) .. controls(-2,-2)  .. (-1.3,-2);

    \node[above, black, font=\small] at (-2.1,-2.7) {Prediction};
    \draw[->, thick, dotted, purple, decoration={markings, mark=at position 0.5 with {\arrow[scale=1.5]{>}}}]
    (-2.8,-2.7) .. controls(-2,-2.7)  .. (-1.3,-2.7);
    
\end{tikzpicture}

        }
    }
    \subfloat[{\scriptsize Incorrect mode}]{\resizebox{0.31\linewidth}{!}{%
            \begin{tikzpicture}
    \fill[gray!30] (-3,-3) rectangle (3,3);
    \fill[white] (-1,-3) rectangle (1,3);
    \fill[white] (-3,-1) rectangle (3,1);

    \pic[scale=0.15] at (-2.5,-0.5) {car_top={red}};
    \pic[scale=0.15, rotate=180] at (-1,0.5) {car_top={darkgray}};
    \pic[scale=0.15, rotate=-270] at (0.5,-2.5) {car_top={green}};




    \draw[->, thick, dotted, purple, decoration={markings, mark=at position 0.5 with {\arrow[scale=1.5]{>}}}]
        (0.5,-2.5) .. controls (0.45,-2)  .. (0.35, -0.2);

    \draw[->, thick, dotted, purple, decoration={markings, mark=at position 0.5 with {\arrow[scale=1.5]{>}}}]
        (-2.5,-0.5) .. controls (-1.5,-0.55)  .. (-0.5,-0.7);

    \draw[->, thick, dotted, purple, decoration={markings, mark=at position 0.5 with {\arrow[scale=1.5]{>}}}]
        (-1,0.5) .. controls(-2,0.55)  .. (-2.8,0.7);

    \node[above, black, font=\small] at (-2.1,-2) {GT};
    \draw[->, thick, blue, decoration={markings, mark=at position 0.5 with {\arrow[scale=1.5]{>}}}]
    (-2.8,-2) .. controls(-2,-2)  .. (-1.3,-2);

    \node[above, black, font=\small] at (-2.1,-2.7) {Prediction};
    \draw[->, thick, dotted, purple, decoration={markings, mark=at position 0.5 with {\arrow[scale=1.5]{>}}}]
    (-2.8,-2.7) .. controls(-2,-2.7)  .. (-1.3,-2.7);
    
\end{tikzpicture}  
        }
    }
    \caption{
        We consider an exemplary intersection scenario, with two interacting vehicles, and one non-interacting vehicle. Both predictions (b) and (c) have similar mean final displacement errors, while only (b) correctly predicts the interaction mode between the green and red vehicle, as in the ground truth (a). 
    }
    \label{fig:example_intersection}
\end{figure}
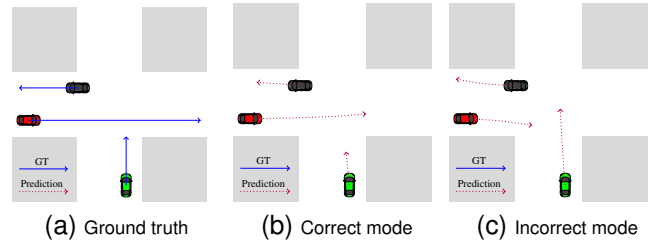

\IEEEpubidadjcol

Vehicle trajectory prediction (VTP) models are evaluated in open-loop, and their performance is primarily evaluated with distance-based metrics that assess the models' accuracy. 
While these metrics are an obvious choice and easy to compute, they heavily depend on datasets, making it impossible to compare models from different datasets and interpret the results. 
Furthermore, none of the existing evaluation frameworks explicitly evaluates the model's ability to correctly predict the correct interaction mode between agents (e.g. the green vehicle lets the red vehicle pass first), which we argue is the most safety-critical aspect of driving.
In \Cref{fig:example_intersection}, we illustrate that traditional distance-based metrics fail to distinguish interactions between agents effectively and that averaging distance errors complicates the interpretation of results.



The aim of this study is to evaluate mode collapse on the interaction level in VTP models in an insightful and more data-independent way. 
More specifically, we want to research when mode collapse occurs and get insights into the temporal dimension of the predictions. 
Towards this end, we introduce a novel evaluation framework to benchmark a model's interaction prediction performance. Our contributions are fourfold:
First, we evaluate the interaction modes of joint trajectory predictions by introducing an explicit metric for mode collapse and utilizing metrics for mode correctness and coverage.
This is a safety-critical aspect that has previously been neglected in VTP evaluation. Second, we introduce time-based variants of these metrics, shedding light on the temporal evolution and consistency of the predictions. 
Third, we filter scenarios and only consider the relevant parts of path-crossing interactions, thereby making the evaluation less dependent on datasets, and improving the interpretability of the metrics. 
Finally, we benchmark two state-of-the-art trajectory prediction models, along with two other baseline models, on the nuScenes dataset and evaluate them using our novel metrics. Our results show that the models suffer from mode collapse and, in some cases, fail to correctly predict the interaction mode between agents, even just before the interaction mode becomes inevitable.



The rest of this paper is organized as follows: In \Cref{sec:related_works} we give a brief literature overview on multimodal trajectory prediction models and the performance metrics used in popular benchmarks. In \Cref{sec:methodology} we present our methodology and formulate our novel metrics. \Cref{sec:models} describes the models that we tested and in \Cref{sec:results} their performance on the nuScenes dataset is discussed, with both qualitative and quantitative results. 
Finally, \Cref{sec:conclusion_discussion} concludes this work, and we discuss limitations as well as exciting directions for future research in this area.

\section{RELATED WORKS}
\label{sec:related_works}

In \Cref{subsec:MTP_modes} we discuss how multimodal trajectory prediction models mitigate mode collapse, what mode representations have been used, and the difference between marginal and joint prediction. 
\Cref{subsec:vtp_metrics} discusses the current trajectory prediction evaluation frameworks, and how they fail to effectively evaluate interactions. 

\subsection{Multimodal trajectory prediction models}
\label{subsec:MTP_modes}
Multimodal trajectory prediction models employ various techniques to mitigate mode collapse. A common approach is to first enumerate diverse possible modes and then condition the prediction upon these modes, to enforce diverse predictions. 
A mode is an abstraction of a trajectory referring to a high-level behavior, and can be represented on the environment level (goal lanes or points) \cite{zhao_tnt_2021}, vehicle level (lane change, accelerating, braking) \cite{berkemeyer_feasible_2021} or interaction level (yielding, going) \cite{kumar_interaction-based_2021}. Using such modes as an intermediate representation to condition the prediction upon, improves interpretability and helps mitigate mode collapse. 
However, since no unified definition of a mode exists, there are also no metrics to quantify the discrete mode prediction performance of the models. 
In this work, we will focus on the interaction modes between agents, and use the concept of free-end homotopy \cite{chen_interactive_2024} to categorize them into clockwise and counterclockwise rotations.

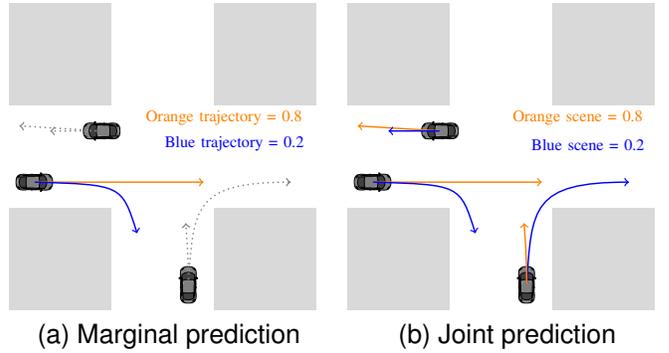
\begin{figure}
    \centering
    \subfloat[Marginal prediction]{
        \resizebox{0.48\linewidth}{!}{%
            \begin{tikzpicture}
    \fill[gray!30] (-3,-3) rectangle (3,3);
    \fill[white] (-1,-3) rectangle (1,3);
    \fill[white] (-3,-1) rectangle (3,1);

    \pic[scale=0.15] at (-2.5,-0.5) {car_top={gray}};
    \pic[scale=0.15, rotate=180] at (-1.2,0.5) {car_top={gray}};
    \pic[scale=0.15, rotate=-270] at (0.5,-2.5) {car_top={gray}};

    \draw[->, thick, gray, dotted, decoration={markings, mark=at position 0.5 with {\arrow[scale=1.5]{>}}}]
        (0.5,-2.5) .. controls (0.6,-1)  and (0.6, -0.5) .. (2.5,-0.5);

    \draw[->, thick, gray, dotted,decoration={markings, mark=at position 0.5 with {\arrow[scale=1.5]{>}}}]
        (0.5,-2.5) .. controls (0.5,-2)  .. (0.45, -1.3);

    \draw[->, thick, orange, decoration={markings, mark=at position 0.5 with {\arrow[scale=1.5]{>}}}]
        (-2.5,-0.5) .. controls (0,-0.5)  .. (0.8,-0.5);

    \draw[->, thick, blue, decoration={markings, mark=at position 0.5 with {\arrow[scale=1.5]{>}}}]
        (-2.5,-0.5) .. controls (-0.8,-0.55)  .. (-0.5,-1.5);

    \draw[->, thick, gray, dotted, decoration={markings, mark=at position 0.5 with {\arrow[scale=1.5]{>}}}]
        (-1.2,0.5) .. controls(-2,0.55)  .. (-2.8,0.6);

    \draw[->, thick,gray, dotted,decoration={markings, mark=at position 0.5 with {\arrow[scale=1.5]{>}}}]
        (-1.2,0.5) .. controls(-2,0.5)  .. (-2.2,0.5);

    \node[above, orange, font=\small] at (1.2,0.5) {Orange trajectory = 0.8};
    \node[above, blue,font=\small] at (1.4,0) {Blue trajectory = 0.2};

\end{tikzpicture}

        }
    }
    \subfloat[Joint prediction]{
        \resizebox{0.48\linewidth}{!}{%
            \begin{tikzpicture}
    \fill[gray!30] (-3,-3) rectangle (3,3);
    \fill[white] (-1,-3) rectangle (1,3);
    \fill[white] (-3,-1) rectangle (3,1);

    \pic[scale=0.15] at (-2.5,-0.5) {car_top={gray}};
    \pic[scale=0.15, rotate=180] at (-1.2,0.5) {car_top={gray}};
    \pic[scale=0.15, rotate=-270] at (0.5,-2.5) {car_top={gray}};

    \draw[->, thick, blue, decoration={markings, mark=at position 0.5 with {\arrow[scale=1.5]{>}}}]
        (0.5,-2.5) .. controls (0.6,-1)  and (0.6, -0.5) .. (2.5,-0.5);

    \draw[->, thick, orange,decoration={markings, mark=at position 0.5 with {\arrow[scale=1.5]{>}}}]
        (0.5,-2.5) .. controls (0.5,-2)  .. (0.45, -1.3);

    \draw[->, thick, orange, decoration={markings, mark=at position 0.5 with {\arrow[scale=1.5]{>}}}]
        (-2.5,-0.5) .. controls (0,-0.5)  .. (0.8,-0.5);

    \draw[->, thick, blue, decoration={markings, mark=at position 0.5 with {\arrow[scale=1.5]{>}}}]
        (-2.5,-0.5) .. controls (-0.8,-0.55)  .. (-0.5,-1.5);

    \draw[->, thick, orange, decoration={markings, mark=at position 0.5 with {\arrow[scale=1.5]{>}}}]
        (-1.2,0.5) .. controls(-2,0.55)  .. (-2.8,0.6);

    \draw[->, thick,blue,decoration={markings, mark=at position 0.5 with {\arrow[scale=1.5]{>}}}]
        (-1.2,0.5) .. controls(-2,0.5)  .. (-2.2,0.5);

    \node[above, orange, font=\small] at (1.5,0.5) {Orange scene = 0.8};
    \node[above, blue,font=\small] at (1.7,0) {Blue scene = 0.2};

\end{tikzpicture}

        }
    }
    \caption{Illustration demonstrating the difference between marginal prediction (a) and joint prediction (b).
    }
    \label{fig:joint_marginal_prediction}
\end{figure}

Multimodal trajectory prediction models can be categorized into node-centric and scene-centric models, which perform the prediction per-agent and jointly for the whole scene, respectively. \Cref{fig:joint_marginal_prediction} demonstrates the difference. 
Generally, scene-centric models better capture the interactions among agents, have higher scene consistency and are more compatible with downstream planners \cite{chen_tree-structured_2023}. On the other hand, node-centric models are easier to train and better cover the agents' motion \cite{luo_jfp_2023}.
In order to evaluate the interaction mode of a trajectory pair, joint trajectory predictions are required. Therefore, we will focus on the evaluation of scene-centric vehicle trajectory prediction models only.

Categorical Traffic Transformer (CTT) \cite{chen_categorical_2023} is an example of such a model. It uses an interpretable set of Scene Modes (SM) to supervise the latent mode distribution. Uniquely, these modes consist of two types: agent2lane (a2l) modes and agent2agent (a2a) modes, thereby capturing both the route and interaction intention of agents. However, because the number of modes scales exponentially with the number of agents in the scene (both in lane options and agent interaction), it is infeasible to predict all modes in scenes with many agents.


\subsection{Trajectory prediction performance metrics}
\label{subsec:vtp_metrics}
Vehicle trajectory prediction models are evaluated in open-loop, using various metrics that assess the accuracy, probability, diversity, and admissibility of the predicted trajectories. 
Distance-based metrics like the minimum average displacement error (minADE), minimum final displacement error (minFDE) and miss rate (MR) have been the primary accuracy metrics used to compare trajectory prediction models. However, the performance of these metrics is heavily dependent on the used dataset, making comparison between different datasets impossible and hindering the interpretation. 

Another aspect that has been neglected in the evaluation is the interactions among agents. Recent works \cite{chen_scept_2022, yuan_agentformer_2021, chen_categorical_2023} have turned to scene-centric models, to better capture interactions between agents by simultaneously rolling-out their future trajectories. The Waymo Open Motion Dataset (WOMD) \cite{ettinger_large_2021} prediction benchmark introduced joint metrics for the minADE, minFDE and MR. Their definitions are similar to their marginal variants, except that the minimum error of $K$ predictions is taken over the whole scene instead of agent-wise. This means that we cannot mix-and-match the best prediction for each agent over different scene samples, which means the prediction task is inherently more challenging but also gives a more realistic idea of the performance. 
While these joint metrics implicitly evaluate agent interactions, the lack of an explicit metric makes interpretation challenging, as demonstrated in \Cref{fig:example_intersection}.

In CTT \cite{chen_categorical_2023} - already introduced above - a2l and a2a modes are defined and used to condition the prediction task upon the scene mode. Additionally, they introduce corresponding mode metrics: the mode correct rate and mode cover rate. The mode correct rate is the percentage of most likely (ML) predictions that match the ground truth (GT) mode (a2a, a2l or both). The mode cover rate is the rate at which one of the $K$ predicted trajectories matches the GT mode.
They compare their performance on these metrics to AgentFormer (AF) \cite{yuan_agentformer_2021} on the nuScenes and WOMD datasets.
While this is a promising step towards formalizing modes and improving intention prediction (lane and interaction modes), their metrics lack interpretability and are still heavily dependent on the dataset. The latter is demonstrated by the fact that for AF there is almost a 50\% performance difference in the a2a cover rate between nuScenes and the WOMD. 
In this work, we build on their mode metrics for a2a interactions, adapt the criteria and extend it to evaluate only critical scenarios and only relevant segments of those scenarios (together with feasible modes), along with evaluating the temporal evolution of the predictions over the scene duration. This allows us to use metrics to quantify a model's interaction prediction performance in a more insightful and data-independent manner.


\section{METHODOLOGY}
\label{sec:methodology}
We argue that current evaluation frameworks lack interpretability because they are constrained to datasets, which vary in size, density, number of agents, etc. Therefore, these frameworks are not able to capture the model's critical interaction prediction performance, because \textit{all} interactions are considered for \textit{all} time-steps. 
We propose to evaluate only the safety-critical interactions, and give a formal definition in \Cref{subsec:interactions}. 
To characterize the interactions, we use a two-class free-end homotopy concept (\Cref{subsec:homotopy}).
Furthermore, we only evaluate the predictions until the point where the interaction class becomes inevitable. To find this point, we simulate feasible future roll-outs for the interacting agents (\Cref{subsesc:rollouts}).
Finally, in \Cref{subsec:metrics} we present our novel metrics for evaluating mode collapse on the interaction level. 
Additionally, we introduce time-based metrics to get insight into the temporal evolution of the predictions by analyzing the consistency of the predictions as the scenario evolves.

\subsection{Filtering safety-critical, interactive scenarios}\label{subsec:interactions}
A unified definition for inter-vehicle interactions was given by 
\cite{markkula_defining_2020}:
\begin{quote}
\textit{``A situation where the behavior of at least two road users can be interpreted as being influenced by the possibility that they are both intending to occupy the same region of space at the same time in the near future."}
\end{quote}

This possibility is very low for a lot of the theoretical number of interactions, as driving is constrained by infrastructure and traffic rules. These interaction pairs are not interesting because e.g. the vehicles are on different lanes in the whole scenario (passing) or they are already in the same lane, (e.g., in car-following scenarios).
The interesting and safety-critical interactions are those where agents initially occupy different lanes but intend to occupy the same region of space or lane in the near future.
Exemplary scenarios are merging and crossings at unsignalized intersections. 

To identify these interaction pairs, we will first formally define Path-Sharing (PS) and then outline our criteria for filtering safety-critical interactions.

\textbf{Path-Sharing (PS).} 
We define the trajectories of two agents as $\tau^{\mathrm{A}}$ and $\tau^{\mathrm{B}}$:
\begin{equation}
\label{eq:traj1}
\tau^{\mathrm{A}} = \left[ (x_1^{\mathrm{A}}, y_1^{\mathrm{A}}), \ldots, (x_N^{\mathrm{A}}, y_N^{\mathrm{A}}) \right]    
\end{equation}
\begin{equation}
\label{eq:traj2}
\tau^{\mathrm{B}} = \left[  (x_1^{\mathrm{B}}, y_1^{\mathrm{B}}), \ldots, (x_N^{\mathrm{B}}, y_N^{\mathrm{B}}) \right]  
\end{equation}
where \( (x_i, y_i) \) is an agent's position at time-step \( t_i \), where \( i = 1, 2, \ldots, N \) and is defined for the maximum interval at which both agents are present in the scene, i.e., recorded in the data.

To determine whether two agents are on the same path, we compute the pairwise distance between both trajectories, from each time-step of $\tau^{\mathrm{A}}$ to all other time-steps of $\tau^{\mathrm{B}}$. 
Thus, the position difference matrix $\Delta P$ is calculated as:
\begin{equation}
\label{eq:dP}
\Delta P = \begin{bmatrix} (x_1^{A},y_1^{A}) \\ \vdots \\ (x_n^{A},y_n^{A})  \end{bmatrix} - \begin{bmatrix} (x_1^{B},y_1^{B}) & \ldots & (x_n^{B},y_n^{B}) \end{bmatrix}
\end{equation}

The resulting matrix is of shape $(n \times n \times 2)$ and the entries $\Delta p_{ij}$ denote the position difference $(\Delta x, \Delta y)_{ij}$ between the agents.
The entries $d_{ij}$ of the distance matrix $D$ (see \cref{eq:d_matrix}) are calculated by taking the Euclidean norm of the position differences \cref{eq:dP}. 
Thus, $d_{ij}$ represents the distance between agent $\mathrm{A}$ at time $t_i$ and agent $\mathrm{B}$ at time $t_j$.
\begin{equation}
\label{eq:d_matrix}
D = \begin{bmatrix} 
    d_{11} & d_{12} & \dots \\
    \vdots & \ddots & \\
    d_{n1} &        & d_{nn} 
    \end{bmatrix}
\qquad
\end{equation}

\begin{figure}[htbp]
    \centering
    \subfloat[Merging]{
        \includegraphics[width=0.45\linewidth]{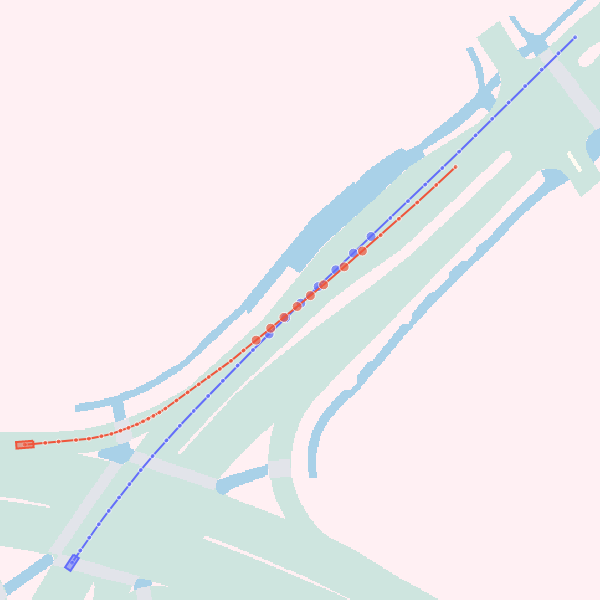}
    }
    \subfloat[Crossing]{
        \includegraphics[width=0.45\linewidth]{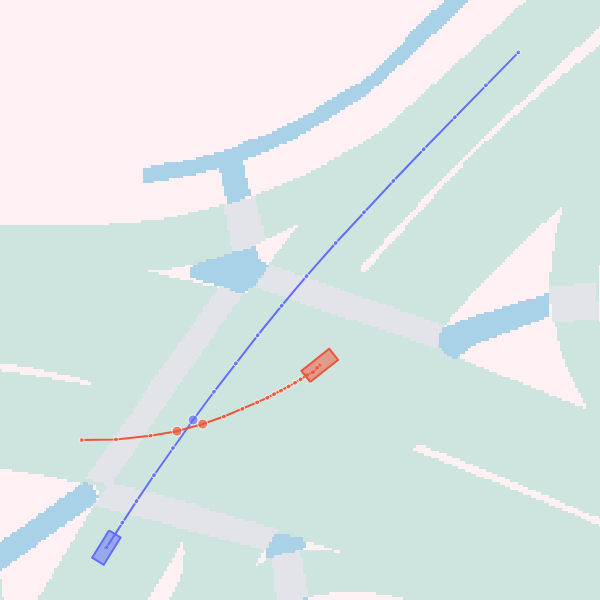}
    }
    \\
    \subfloat[Car-following]{
        \includegraphics[width=0.45\linewidth]{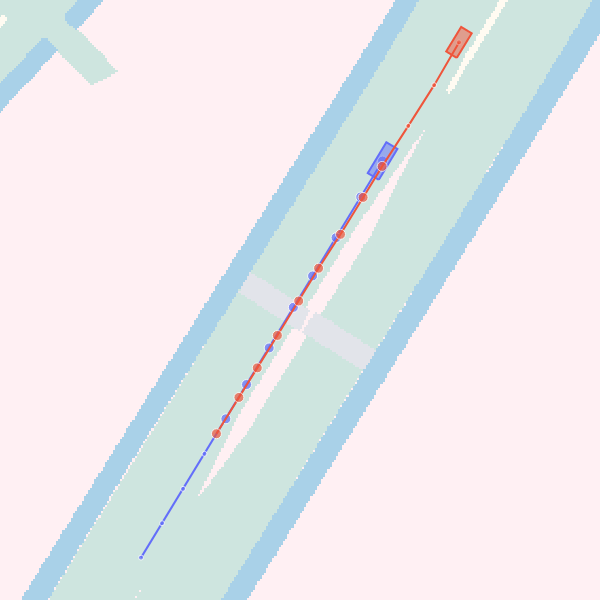}
    }
    \subfloat[Passing vehicles]{
        \includegraphics[width=0.45\linewidth]{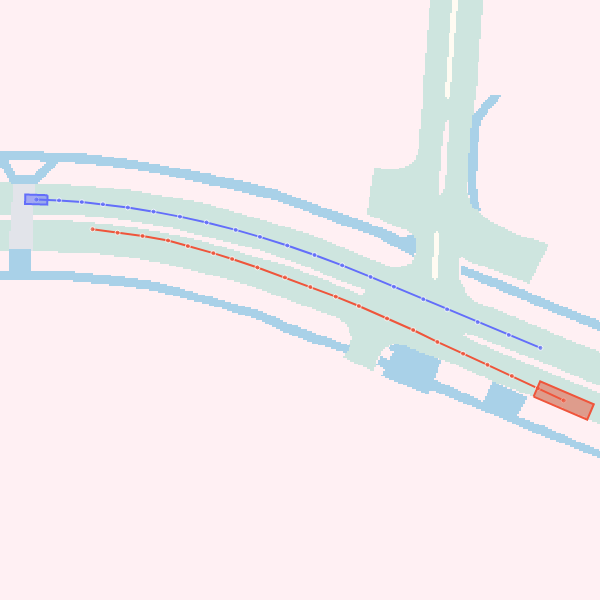}
    }
    \caption{Exemplary traffic scenarios of safety-critical interaction agent-pairs (a, b) and non-interacting agent-pairs (c, d) from the nuScenes dataset. The time-steps where the agents are on the commonly shared path are visualized with big markers ($i_\mathrm{PS}$ time-steps).
    }
    \label{fig:examples_interaction}
\end{figure}

A path point is on the commonly shared path if the distance is smaller than a threshold, $d_{ij} < d_{\mathrm{collision}}$.
Thus, to determine the set of time-steps for which the other agent occupies the same path, we calculate the PS vector for each agent-pair as:
\begin{equation}
\label{eq:onpath1}
\mathbf{i}_\mathrm{PS}^{\mathrm{A}} = \{i \mid \exists j, \; d_{ij} < d_{\mathrm{collision}} 
\}
\end{equation}
\begin{equation}
\label{eq:onpath2}
\mathbf{i}_\mathrm{PS}^{\mathrm{B}} = \{j \mid \exists i, \; d_{ij} < d_{\mathrm{collision}} \}
\end{equation}
In \Cref{fig:examples_interaction}, we show exemplary scenarios of interacting and non-interacting agent-pairs, where the $i_\mathrm{PS}$ time-steps are visualized with bigger markers. 
To make sure the real minimum distance is calculated, the position vectors can be interpolated to increase the precision for the distance calculation. 

\textbf{Interaction criteria. }
We define an interaction pair to be safety-critical if the trajectories are not PS at first, but are PS at a later stage. We define the time at which an agent starts to be on the commonly shared path based on the first time-step ($i_\mathrm{PS}=\min(\mathbf{i}_\mathrm{PS})$) as:
\begin{equation}
t_{\mathrm{PS}} = t(i_\mathrm{PS})
\end{equation}
Thus, we are looking for the interactions where both of the following holds:
\begin{equation}
\label{eq:pathsharing_A}
t_{\mathrm{PS}}^{\mathrm{A}} > t_{\mathrm{min}} 
\end{equation}
\begin{equation}
\label{eq:pathsharing_B}
t_{\mathrm{PS}}^{\mathrm{B}}  > t_{\mathrm{min}}
\end{equation}
To characterize the closeness of the interactions, we also calculate the time difference between the instances at which the vehicles begin to occupy the shared region:
\begin{equation}
\label{dt_def}
   \Delta t_{\mathrm{PS}} =| t_{\mathrm{PS}}^{\mathrm{B}} - t_{\mathrm{PS}}^{\mathrm{A}} | 
\end{equation}
If the sequence is long, two trajectories can be PS at the end, even if the cars are very far apart. Therefore, we impose an additional time-based constraint on the interaction: 
the time difference $\Delta t_{\mathrm{PS}}$ should not be more than a threshold $\Delta t_{\mathrm{PS, max}}$:
\begin{equation}
\label{dt_constraint}
   \Delta t_{\mathrm{PS}} \leq \Delta t_{\mathrm{PS, max}}
\end{equation}
In \cref{subsec:experimental_setup} we define the thresholds for our experiments and in \cref{subsec:nuscenes_stats} we will analyze the closeness of the interactions in the nuScenes dataset.


%
Finally, an interaction is defined to be safety-critical, if the trajectory pair satisfies all three conditions, i.e. \cref{eq:pathsharing_A}, \cref{eq:pathsharing_B}, and \cref{dt_constraint}.
With this definition, we can filter the safety-critical interesting interactions, like merging and crossing, from basic car-following and traffic light scenarios. 
This reduces the dependency on the dataset, as we only evaluate similar and safety-critical interactions.
In the traditional trajectory prediction evaluation, all cars and scenes are considered, which complicates interpretation, because the distance errors are averaged, making it unclear what kind of scenarios were evaluated and how the model performed in critical cases.
Thus, by applying our methodology, the metrics become more interpretable and insightful.

\subsection{Categorizing interaction mode using homotopy}\label{subsec:homotopy}
To categorize the interaction mode between agent-pairs, we use the concept of homotopy classes.
A group of trajectories with common start- and endpoints belong to the same homotopy class if they can be continuously deformed into each other without intersecting any obstacle \cite{bhattacharya_topological_2012}. 
We will follow \cite{chen_interactive_2024} and build upon their concept of free-end homotopy, which has more flexible classes because the end-point of the trajectories does not have to be shared, which is useful for multiple prediction trajectories.
The agents' interaction class is determined by the winding angle, which is the cumulative angular difference between the agent-pair. 
Let $\mathbf{\tau}^{{A}}$ be the trajectory of agent A and $\mathbf{\tau}^{{B}}$ be the trajectory of agent B, with the sequence of waypoints discretized as $\left[\left(x_i^{\mathrm{A}}, y_i^{\mathrm{A}}\right)\right]_{i=1}^N$ and $\left[\left(x_i^{\mathrm{B}}, y_i^{\mathrm{B}}\right)\right]_{i=1}^N$. The angular distance $\Delta \theta \left(\mathbf{\tau}^{{A}}, \mathbf{\tau}^{{B}}\right)$ is computed as:
\begin{equation}
\label{angular_distance}
    \sum_{i=1}^{N-1} \arctan \frac{y_{i+1}^\mathrm{A}-y_{i+1}^{\mathrm{B}}}{x_{i+1}^\mathrm{A}-x_{i+1}^{\mathrm{B}}}-\arctan \frac{y_i^\mathrm{A}-y_i^{\mathrm{B}}}{x_i^\mathrm{A}-x_i^{\mathrm{B}}}.
\end{equation}
\begin{figure}
    \centering
    \subfloat[CW class]{
        \resizebox{0.48\linewidth}{!}{%
            \begin{tikzpicture}
    \fill[gray!30] (-3,-3) rectangle (3,3);
    \fill[white] (-1,-3) rectangle (1,3);
    \fill[white] (-3,-1) rectangle (3,1);

    \pic[scale=0.15] at (-2.5,-0.5) {car_top={red}};
    \pic[scale=0.15] at (-2.5+0.33,-0.5) {car_top={red}};
    \pic[scale=0.15] at (-2.5+0.66,-0.5) {car_top={red}};
    \pic[scale=0.15] at (-1.5,-0.5) {car_top={red}};
    
    \pic[scale=0.15, rotate=-90] at (-0.5,1.5) {car_top={orange}};
    \pic[scale=0.15, rotate=-90] at (-0.5,0.5) {car_top={orange}};
    \pic[scale=0.15, rotate=-90] at (-0.5,-0.5) {car_top={orange}};
    \pic[scale=0.15, rotate=-90] at (-0.5,-1.5) {car_top={orange}};

    \foreach \i in {0,1,2,3} {
        \pgfmathsetmacro{\x}{-2.5 + 0.33*\i}
        \pgfmathsetmacro{\y}{1.5 - \i}
        \draw[->, thick, green] (\x,-0.5) -- (-0.5,\y);
    }
\end{tikzpicture}

        }
    }
    \subfloat[CCW class]{
        \resizebox{0.48\linewidth}{!}{%
            \begin{tikzpicture}
    \fill[gray!30] (-3,-3) rectangle (3,3);
    \fill[white] (-1,-3) rectangle (1,3);
    \fill[white] (-3,-1) rectangle (3,1);

    \pic[scale=0.15] at (-2.5,-0.5) {car_top={red}};
    \pic[scale=0.15] at (-2.5+1,-0.5) {car_top={red}};
    \pic[scale=0.15] at (-2.5+2,-0.5) {car_top={red}};
    \pic[scale=0.15] at (-2.5+3,-0.5) {car_top={red}};
    
    \pic[scale=0.15, rotate=-90] at (-0.5,1.5) {car_top={orange}};
    \pic[scale=0.15, rotate=-90] at (-0.5,1.5-0.33) {car_top={orange}};
    \pic[scale=0.15, rotate=-90] at (-0.5,1.5-0.66) {car_top={orange}};
    \pic[scale=0.15, rotate=-90] at (-0.5,1.5-1) {car_top={orange}};

    \foreach \i in {0,1,2,3} {
        \pgfmathsetmacro{\x}{-2.5 + 1*\i}
        \pgfmathsetmacro{\y}{1.5 - 0.33*\i}
        \draw[->, thick, green] (\x,-0.5) -- (-0.5,\y);
    }
\end{tikzpicture}

        }
    }
    
    \caption{Visualization of angular distance calculation and convergence of two agents traversing an intersection. The figure shows the cars and their relative angles at four consecutive time-steps. In (a) the red vehicle yields, resulting in a clockwise rotation, and in (b) the orange vehicle yields, resulting in a counterclockwise rotation.
    }
    \label{fig:winding_number_convergence}
\end{figure}
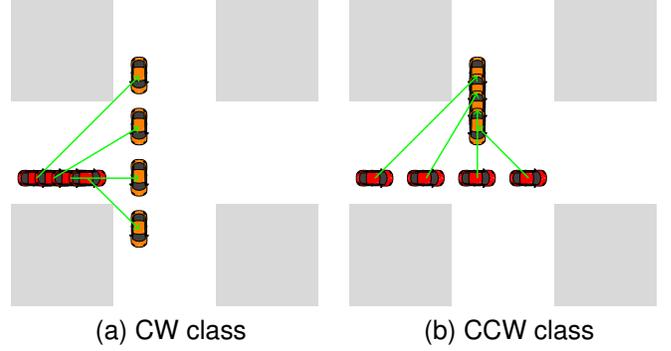
The angle describes the relative rotation of the agents with respect to each other, as illustrated by \Cref{fig:winding_number_convergence}. The angle is used to categorize the agent-to-agent (a2a) interaction into three modes: $[S$, $C W, C C W]$ (static, clockwise, counterclockwise):
\begin{equation}
\label{homotopy_classes}
\mathrm{h}:=\left\{\begin{array}{rr}
C W, & \Delta \theta\left(\mathbf{\tau}^{\mathrm{A}}, \mathbf{\tau}^{\mathrm{B}}\right)<-\hat{\theta} \\
S, & -\hat{\theta} \leq \Delta \theta\left(\mathbf{\tau}^{\mathrm{A}}, \mathbf{\tau}^{\mathrm{B}}\right) < \hat{\theta} \\
C C W, & \Delta \theta\left(\mathbf{\tau}^{\mathrm{A}}, \mathbf{\tau}^{\mathrm{B}}\right) \geq \hat{\theta}.
\end{array}\right.
\end{equation}
Where $\hat{\theta}$ is a fixed threshold, large enough to differentiate between the classes. 
The $CW$ and $CCW$ classes are visualized in \Cref{fig:winding_number_convergence}. The $S$ class emerges when the relative rotation between the agents is very small, e.g., in car-following scenarios or when the agents are very far apart. 



In contrast to \cite{chen_categorical_2023, chen_interactive_2024}, we set the threshold $\hat{\theta}$ to zero, effectively eliminating the static class. 
A fixed non-zero threshold can lead to ambiguities, as the angular distance $\Delta \theta$ not only depends on the speed and intention of the agents, but also on the road topology and the used prediction horizon. By eliminating the static class, we always have a distinct interaction class for a trajectory pair.
This is especially important for the predictions, as they might not be close to the ground truth, but still contain the model's implicit homotopy class prediction, i.e., the intuition for how the agents will interact (CW or CCW rotation with respect to each other). 
Furthermore, as described in \Cref{subsec:interactions}, we will only evaluate the safety-critical path-crossing interactions. Therefore, most static interactions like car-following and distant agent-pairs will be already filtered out, making the static class redundant anyway. 

Besides filtering interaction scenarios for evaluation, we also want to segment the interesting temporal duration of those scenarios, i.e., once an interaction has happened, there is no point in further evaluating it.
E.g. \Cref{fig:winding_number_convergence} depicts the process of calculating the angular distance over time for two agents traversing an intersection. From this figure, it becomes clear that the angular distance is only significant if the agents are close. Furthermore, once either of the agents has entered the shared path (the middle of the intersection in this case), the homotopy class of the interaction is inevitable and the angular distance converges afterward. Geometrically, this happens once $d_{ij} < d_{\mathrm{collision}}$ for either of the agents. 
However, it would be too imprecise to use this as criterion, as the real moment is earlier, because vehicles cannot instantly accelerate and decelerate.
Thus, to find the true instance at which the homotopy class becomes inevitable, dynamic simulations are needed, which will be discussed in the next subsection.

\subsection{Enumerating feasible homotopy classes}\label{subsesc:rollouts}
In this subsection, we describe our methodology for enumerating potential homotopy classes that are feasible in a given scenario by simulating feasible future trajectories for agent-pairs.
Our goal is to find the set of feasible homotopy classes and the true Inevitable Homotopy State (IHS) - a state after which only one homotopy class is feasible. These are important concepts for the novel metrics we propose in \Cref{subsec:metrics}. 
For each time-step in the scene, and all agent-pairs that will cross paths in the near future, we want to find the set of feasible homotopy classes.
To find this set, we accelerate one agent and decelerate the other, and vice versa. Thus, the set of future roll-outs for agent-pair $(A,B)$ at time-step $t$ is: 
\begin{equation}
    \mathbf{y}_\mathrm{roll,t} = [(\tau_\mathrm{decel}^\mathrm{A}, \tau_\mathrm{accel}^\mathrm{B}), (\tau_\mathrm{accel}^\mathrm{A}, \tau_\mathrm{decel}^\mathrm{B}) ].
\end{equation} 
We keep the ground truth paths of the agents, and manipulate only the velocity profile of the agents (either accelerate or decelerate), while keeping both longitudinal and lateral accelerations within realistic limits for comfortable driving. 

Finally, we check whether a roll-out pair is feasible by using a binary collision detection function, denoted by $\operatorname{IsCollision}(\tau^\mathrm{A}, \tau^\mathrm{B})$. To take into account the vehicle dimensions and headings, we take inspiration from \cite{ziegler_fast_2010}, and fit three disks with radii $r^\mathrm{A} = \frac{1}{2}\mathrm{width}$ of to each vehicle: at the vehicle center and at both bumpers. A collision is detected by computing the minimum distances between all disks of both vehicles, for all time-steps. The vehicles are in collision if the minimum distance $d_{\mathrm{min}}$ is smaller than the sum of the disk radii $r$ fitted to the vehicles: $d_{\mathrm{min}} < r^\mathrm{A} + r^\mathrm{B}$. 
Whilst there can still be hypothetical cases where a collision is missed, this approach works in most practical cases and is computationally efficient.

\begin{figure*}[hbt!]
\centering
\includegraphics[width=\textwidth]{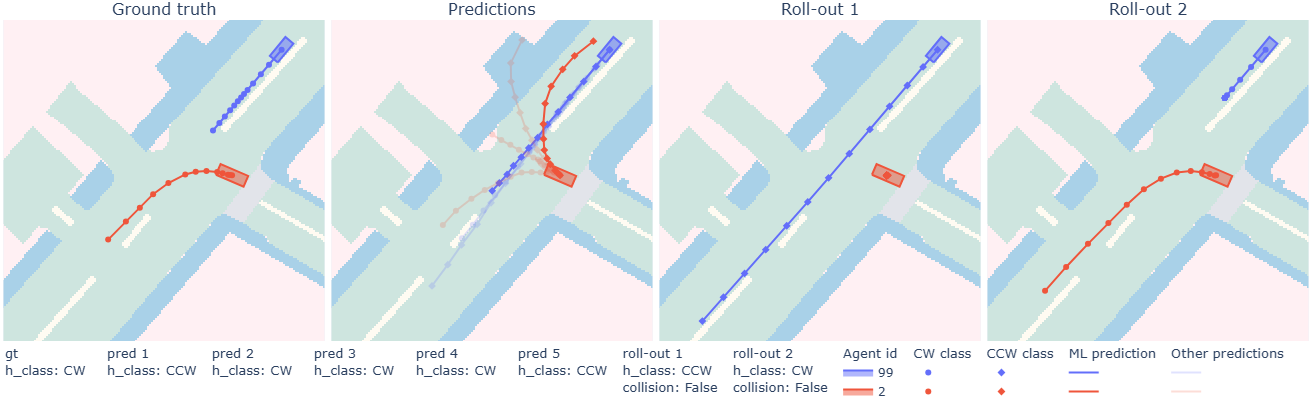}
\caption{Visualization of the interaction mode evaluation for AgentFormer on agent-pair (99,2) at frame 11 in scene-0103 of the nuScenes dataset. For the predictions, only the ML prediction is shown with full opacity. The corresponding homotopy classes ($\mathrm{h\_class}$) are shown in the legend, and displayed in the plot using $\circ$ and $\diamond$ markers for the CW and CCW class, respectively. Additionally, the collision status is shown in the legend for both roll-outs. For this specific frame, both interaction modes are still feasible. The mode is not predicted correctly (ML prediction), but it is covered by one of the other predictions.} 
\label{fig:example_vis_method}
\end{figure*}

Now, we can define the set of feasible roll-outs as:
\begin{equation}
\mathbf{y}_\mathrm{feas,t}  = \{ (\tau^\mathrm{A}, \tau^\mathrm{B}) \in \mathbf{y}_\mathrm{roll,t} \mid \neg \operatorname{IsCollision}(\tau^\mathrm{A}, \tau^\mathrm{B}) \} .
\end{equation}
And consequently, the unique set of feasible homotopy classes is:
\begin{equation}
\mathbf{h}_\mathrm{feas,t} = \{h(\tau^\mathrm{A}, \tau^\mathrm{B}) \mid (\tau^\mathrm{A}, \tau^\mathrm{B}) \in  \mathbf{y}_\mathrm{feas,t} \},
\end{equation}
where $h \in \{CW, CCW\}$. 
Thus, we define the \textbf{Inevitable Homotopy State (IHS)}, as the point in time at which only one unique homotopy class is feasible (non-colliding):
\begin{equation}
t_\mathrm{h,collapse} = \min \{\ t \ \big| \ \ |\mathbf{h}_{\mathrm{feas,t}}| = 1 \ \}.\\
\end{equation}


\subsection{Evaluating interaction mode prediction performance}\label{subsec:metrics} 

Since we want to evaluate the interaction between trajectories of agent-pairs, we require joint multi-agent predictions. Vehicle trajectory prediction models are multimodal, meaning they predict a set of $K$ diverse trajectories, with corresponding probabilities. 
Let us denote the predicted modalities of agent-pair $(\mathrm{A},\mathrm{B})$ as $\mathbf{y}_\mathrm{pred}^{\mathrm{A},\mathrm{B}} = \{(\tau^\mathrm{A}, \tau^\mathrm{B})_{1}, \hdots, (\tau^\mathrm{A}, \tau^\mathrm{B})_{K} \}$,  with the predictions ordered in decreasing likelihood, so $(\tau^\mathrm{A}, \tau^\mathrm{B})_{1}$ corresponds to the most likely (ML) prediction. The set of homotopy classes of the model's predictions is:
\begin{equation}
\mathbf{h}_\mathrm{pred} = \{h((\tau^\mathrm{A}, \tau^\mathrm{B})_\mathrm{k}) \mid (\tau^\mathrm{A}, \tau^\mathrm{B})_\mathrm{k}) \in  \mathbf{y}_\mathrm{pred} \}.
\end{equation}
Furthermore, we denote $h_\mathrm{ml}$ the homotopy class of the ML prediction, and $h_\mathrm{gt}$ the ground truth homotopy class.
To evaluate the model's ability to correctly predict the interaction mode, we follow \cite{chen_categorical_2023} in defining mode correctness and coverage. The a2a \textbf{mode is correct} if the ML prediction's mode corresponds to the ground truth mode, i.e., $h_\mathrm{ml} = h_\mathrm{gt} $. The a2a \textbf{mode is covered} if one of the $K$ predictions covers the ground truth mode, i.e., $h_\mathrm{gt} \in \mathbf{h}_\mathrm{pred} $.

In contrast to the default setting in VTP evaluation, we will not evaluate these metrics for the whole scene. Instead, we only consider the safety-critical interactions, and only till the inevitable homotopy state, denoted by $h_{\mathrm{gt,final}}$.
Thus, we evaluate till the last point at which both classes are still feasible, i.e.:
\begin{equation}
t_\mathrm{h,final} = \max \{\ t \ \big| \ \ |\mathbf{h}_\mathrm{feas, t}| = 2 \ \}.\\
\end{equation}

The evaluation starts once the homotopy class starts to converge towards the inevitable homotopy state, but at most a whole prediction horizon $T_{\mathrm{p}}$ before then:
\begin{equation}
\begin{split}
&t_{\mathrm{h,start}} = \min  \{\ \ t \ \big| \ \ h_{\mathrm{gt,t}} = h_{\mathrm{gt,final}})\ \} \quad \\
&\text{for} \quad t \in [t_{\mathrm{h,final}} - T_{\mathrm{p}}, t_{\mathrm{h,final}}].
\end{split}
\end{equation}
Thus the evaluation interval is $[t_\mathrm{h,start}, t_\mathrm{h,final}]$.
Note that the duration of this interval varies, and in many cases is shorter than the 6-second prediction horizon, because the interval for which both agents are recorded in the data is shorter or the ground truth homotopy class starts to converge later.
To get insight into the temporal evolution of the interaction prediction, we propose a time-based metric: the time-to-correct-mode-prediction ($\Delta T_{\mathrm{correct}}$), which is the time the model needs to recognize the intention of the cars before the interaction has settled, i.e., before the inevitable homotopy state is reached.
\begin{equation}
\begin{split}
&t_{\mathrm{incorrect}} = \max \{t \mid h_\mathrm{gt,t} \neq h_\mathrm{ml,t}\}\\
&\Delta T_{\mathrm{correct}} = t_\mathrm{h,final} - t_{\mathrm{incorrect}}
\end{split}
\end{equation}
Similarly, we compute the time-to-covered-mode-prediction ($\Delta T_{\mathrm{covered}}$). The difference is that we consider all $K$ predictions of the model, instead of just the most likely one. 
\begin{equation}
\begin{split}
&t_{\mathrm{uncovered}} = \max \{t \mid h_\mathrm{gt,t} \not\in \mathbf{h}_\mathrm{pred,t}\}\\
&\Delta T_{\mathrm{covered}} = t_\mathrm{h,final} - t_{\mathrm{uncovered}}
\end{split}
\end{equation}
If the predictions are correct or covered from the beginning of the prediction interval, we cannot calculate the respective times, because we cannot make any assumptions about the model's predictions before then. In these cases, we consider the predictions a discrete correct class rather than a time. Therefore, we report three metrics, aggregated over all interactions: the percentage of predictions that are correct or covered from the beginning of the interaction interval ($@ T_{\mathrm{pred}}$), the mean times for the predictions that are not, and the percentage of predictions where the $\Delta T = 0$, meaning the predictions are wrong even just before the inevitable homotopy collapse state~($@0s$).

\textbf{Mode collapse.} 
We define a2a mode collapse as an interaction mode being feasible, but not predicted by any of the model's predictions, i.e., $\mathbf{h}_\mathrm{feas} \not\subseteq \mathbf{h}_\mathrm{pred}$. 
So, mode collapse does not necessarily consider the ground truth, but the feasibility of hypothetical future roll-outs. 
Finally, we define the mode collapse rate as the percentage of time-steps in the relevant interval $t \in [t_\mathrm{h,start}, t_\mathrm{h,final}]$, where mode collapse occurs. 
It is worth noting that in many cases (i.e., scenes with many agents) it is impossible for the model to cover all feasible modes with a finite number of joint predictions, due to the cardinality of the mode space growing exponentially with the number of agents. 

\textbf{Temporal consistency of predictions.} In order to plan a safe motion, the model's predictions should stay somewhat consistent throughout the scene, i.e., small variations in motions in consecutive time-steps should not cause prediction of a new mode. Therefore, we propose to also evaluate the consistency of the ML prediction's interaction mode. The prediction consistency is a hit-or-miss metric, evaluated for each pair within the aforementioned relevant time horizon. The predictions for an agent-pair are said to be \textbf{consistent} if the model's ML mode prediction changes at most one time. E.g., given the mode predictions of consecutive time-steps are $[CW, CCW, CCW]$, the predictions are said to be consistent, as it is acceptable for the model to correct itself. On the other hand, consecutive mode predictions of $[CCW, CW, CCW]$ are considered to be inconsistent. 
The temporal consistency of the predictions was found to be an important factor for the planner's performance in closed-loop simulation in \cite{chen_tree-structured_2023}.


\begin{table}[t]
    \centering
    \caption{Example mode metrics evolution for AF's predictions on agent-pair (99,2) in scene-0103 of the nuScenes dataset.
    The ground truth mode is covered from the beginning, but only predicted consistently correct from frame 13 onwards. Thus, the predictions are inconsistent in this case.
    }
    \label{tab:example_tab_method}
    \resizebox{\columnwidth}{!}{
    \begin{tabular}{cccccccc}
        \toprule
        frame & \shortstack{GT \\ mode} & \shortstack{ML \\ mode} & \shortstack{all K \\ modes} & \shortstack{feasible \\ modes} & \shortstack{mode \\ correct} & \shortstack{mode \\ covered} & \shortstack{mode \\ collapse} \\
        \midrule
        5 & CW & CW & CW & CCW, CW & \checkmark & \checkmark & \checkmark \\
        6 & CW & CW & CW & CCW, CW & \checkmark & \checkmark & \checkmark \\
        7 & CW & CW & CW & CCW, CW & \checkmark & \checkmark & \checkmark \\
        8 & CW & CW & CW & CCW, CW & \checkmark & \checkmark & \checkmark \\
        9 & CW & CW & CW & CCW, CW & \checkmark & \checkmark & \checkmark \\
        10 & CW & CW & CW & CCW, CW & \checkmark & \checkmark & \checkmark \\
        11 & CW & CCW & CCW CW & CCW, CW &  & \checkmark &  \\
        12 & CW & CCW & CCW CW & CCW, CW &  & \checkmark & \\
        13 & CW & CW & CW & CCW, CW & \checkmark & \checkmark & \checkmark \\
        14 & CW & CW & CW & CCW, CW & \checkmark & \checkmark & \checkmark \\
        15 & CW & CW & CW & CCW, CW & \checkmark & \checkmark & \checkmark \\
        16 & CW & CW & CW & CW & \checkmark & \checkmark &  \\
        \bottomrule
    \end{tabular}}
\end{table}

\textit{Implementation example.}
Let us look at an example from AgentFormer's (AF) \cite{yuan_agentformer_2021} predictions on one of the validation scenes of the nuScenes dataset. In this scene, only the relevant interacting agent-pairs are considered. At each frame, we simulate future roll-outs and check their feasibility with the collision checker. Furthermore, we compute the homotopy classes of the ground truth, the predictions and the roll-outs. In \Cref{fig:example_vis_method}, we visualize this process for a single frame.
\Cref{tab:example_tab_method} shows an overview of the interaction modes of predictions and roll-outs for all relevant frames of this interaction-pair.

For this specific interaction, the inevitable homotopy state is at frame 16, as only the $CW$ mode is still feasible from then, as the $CCW$ mode would end in a collision. We wish to evaluate the mode predictions for a whole prediction horizon $T_\mathrm{p}$ before then. However, in many cases (such as this example), this is not possible, simply because the interval for which both agents are recorded in the dataset is not long enough. Thus, we will evaluate the mode predictions from the first point at which there are predictions for both agents, until the inevitable homotopy state. In this case: from frame 5 until frame 15. Since nuScenes is recorded at $2 \mathrm{Hz}$, we find that it takes the model $\Delta T_{\mathrm{correct}} = 1.5 \mathrm{s}$ to correctly predict the interaction class.
For $\Delta T_{\mathrm{covered}}$ we see that the predictions cover the ground truth class from the start of the interaction interval. Since there are no prior time-steps available, we consider such cases a correct/covered class, rather than a time.
Furthermore, from the table, it becomes clear that the predictions are inconsistent because the ML prediction's mode changes more than once. 
Finally, in 9 out of the 11 frames the model did not predict all feasible modes, so the mode collapse rate for this scene is $81.8 \%$.

\section{TRAJECTORY PREDICTION MODELS}
\label{sec:models}

We test our novel evaluation methodology on the nuScenes dataset \cite{caesar_nuscenes_2020} and report results for four models: AgentFormer (\Cref{subsec:AF}), Categorical Traffic Transformer (\Cref{subsec:CTT}), a constant velocity model (\Cref{subsec:CV}) and an oracle model (\Cref{subsec:oracle}).
In the following subsections, we briefly discuss the characteristics and implementation of these models.




\subsection{AgentFormer}
\label{subsec:AF}
AgentFormer (AF) \cite{yuan_agentformer_2021} is a multi-agent trajectory prediction model.
They utilize a transformer-based architecture, that simultaneously models the social and temporal dimension of agents. 
Their prediction framework jointly models the agents' intentions, to predict diverse and socially-aware future trajectories\cite{yuan_khrylxagentformer_2024}. 
We will utilize their pre-trained nuScenes model, and use the version which outputs $K=5$ multi-agent trajectories.

\subsection{Categorical Traffic Transformer}
\label{subsec:CTT}
Categorical Traffic Transformer (CTT) \cite{chen_categorical_2023} is a multi-agent trajectory prediction model, with an interpretable latent space consisting of agent-to-agent and agent-to-lane modes. CTT generates diverse behaviors by conditioning the trajectory prediction on different modes. 
The authors published their code including pre-trained weights for the nuScenes dataset \cite{noauthor_nvlabsdiffstack_nodate}.
Unfortunately, we did not succeed in reproducing the numbers reported in their paper and uncovered various issues, making direct comparison with the other models difficult.
Firstly, their pre-trained model is trained for a prediction horizon of 3 seconds, whereas AF is trained for 6 seconds, as dictated by the nuScenes benchmark. 
To match the varying prediction horizons, the 6-second predictions from AF are cut to 3 seconds. 
Secondly, all $K$ predicted modes and trajectories are identical, making the model effectively unimodal. 
Finally, whereas AF predicts for all vehicles in the scenes, CTT predicts only for the road users within a certain attention radius of the ego-vehicle, but it does include pedestrians whereas AF does not.
We use AF's data preprocessing backbone and match CTT's predictions to the corresponding agents. However, due to the aforementioned attention radius used in CTT, many predictions are missing for certain agents. In these cases, the current ground truth position is kept static and used as a prediction instead.
Due to these issues, we are not able to report the real performance of CTT on interaction prediction. However, we still report the metrics and compare them to the other models, to set a baseline and show that our methodology generalizes to other models.




\subsection{Constant velocity model}
\label{subsec:CV}
The constant velocity (CV) model is a simplistic unimodal model that assumes the vehicle will remain in its current heading and velocity \cite{karle_scenario_2022}. Because it produces a single mode, it inherently suffers from mode collapse. 
However, it is an interesting baseline for comparison, because it tells us in how many scenarios we can correctly assess the vehicle pair's interaction class by simply extrapolating their current trajectories.

\subsection{Oracle model}
\label{subsec:oracle}
As the cardinality of the space of interaction modes grows exponentially with the number of agents in the scene, and trajectory prediction models typically predict a fixed set of $K$ modes, covering all feasible modes becomes infeasible in scenes with many agents.
To test this limitation, we propose a multimodal oracle model. The oracle's goal is to predict a set of $K$ multimodal trajectories that cover all feasible modes of the interacting agents. 
The oracle will be given access to the agents' ground truth paths, so it knows which agents will be interacting, i.e., crossing the same path, in the near future. However, the trajectories are unknown, i.e., it does not know the velocity profiles along the path, so the interaction class is still to be determined by the model. The oracle's goal is to cover all feasible interaction modes between the path-crossing agent-pairs. 
Analogously to the methodology described in \Cref{subsesc:rollouts}, we keep the agents' ground truth paths and simulate future roll-outs with a constant velocity, deceleration, or acceleration profile. 
Firstly, all agents are initialized with their constant velocity profile. 
Next, we determine all combinations of constant velocity, acceleration, and deceleration profiles between the interacting agents and reject the combinations with collisions. 
Finally, we must assign each joint prediction a likelihood. We argue that the likelihood of a joint scene prediction is proportional to the overall utility in the scene, where the average speed of a roll-out combination can be used as a utility measure. Therefore, to get a finite set of $K$ joint predictions, we compute the average speed of the roll-outs and output the top-$K$ trajectory combinations with the highest average speed. 

\section{RESULTS}
\label{sec:results}
Our aim is to evaluate the interaction mode prediction performance of VTP models in an insightful and data-independent way. 
More specifically, we want to research when mode collapse happens and get insight into the temporal dimension of the predictions.  
First, we define the parameters used in our experiments in \Cref{subsec:experimental_setup}.
Second, we employ our methodology for finding path-crossing safety-critical interactions on the widely used nuScenes traffic dataset, and report interaction statistics in \Cref{subsec:nuscenes_stats}. 
Next, we test four baseline models (described in \Cref{sec:models}) and evaluate their performance using our novel evaluation framework in \Cref{subsec:interaction_metrics_results}.
We show that mode collapse happens and provide insights into the temporal evolution of the predictions.
Finally, compare our metrics to the traditional distance-based metrics in \Cref{subsec:distance_metrics}.

\subsection{Experimental setup}
\label{subsec:experimental_setup}
This section discusses the hyperparameters of our evaluation framework\footnote{\label{fn:code}Our code is available online at \\ \href{https://github.com/MaartenHugenholtz/InteractionEval}{https://github.com/MaartenHugenholtz/InteractionEval}}, to find the safety-critical interactions in a dataset, simulate feasible roll-outs, and test models on our interaction metrics. 

First, the $\mathrm{PS}$ vectors are computed for all possible agent-pairs, to find the time steps where the agents are on the shared path. 
We empirically found $d_{\mathrm{collision}} = 1.5 \,\mathrm{m}$ to be a reasonable threshold, considering that two narrow cars would be in collision if the distance between their path centerlines is less than 1.5\,m. 
Next, we need to filter out the path-sharing interactions, where there is a big time difference between the agents starting to occupy the same path.
After carefully considering various scenarios from the nuScenes dataset, we found $\Delta t_{\mathrm{PS, max}} = 6 \, \mathrm{s} $ to be an appropriate threshold.

For simulating the roll-outs, we need to respect acceleration and velocity limits.
The absolute longitudinal acceleration limit is set to $|a_{\mathrm{lon}}| \leq 1.47\,  \mathrm{m/s^2}$ and the lateral to $|a_{\mathrm{lat}}| \leq 1.18\,  \mathrm{m/s^2}$, which is based on the values from \cite{de_winkel_standards_2023}. 
For the accelerations, we also set the maximum velocity equal to the maximum velocity of the scene, thereby implicitly respecting any speed limits or traffic that influences the maximum velocity in the scene.

\subsection{Interaction statistics nuScenes}
\label{subsec:nuscenes_stats}
\begin{figure}
    \centering
    \includegraphics[width=0.45\textwidth]{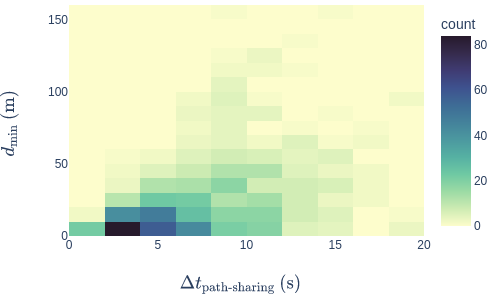}
    \caption{Density heatmap of the path-sharing interactions in the full train-validation split of nuScenes. The interactions are characterized in closeness, with the real-time closest distance on the y-axis and the time difference between the agents occupying the shared-path on the x-axis. }
    \label{fig:path_sharing_density}
\end{figure}

We analyzed interactions across the entire train and validation splits of the nuScenes dataset, and applied our methodology to identify safety-critical interactions. 
In total, we identified 18,299 theoretical interactions across the entire dataset. The theoretical upper limit per scene is calculated as \(N(N-1)/2\), considering the symmetry of interactions and the absence of self-pairs. 
However, in reality, only 16,756 theoretical interaction pairs exist, as not all agents are recorded for the full scene duration.
After applying our first two interaction criteria, i.e., the agents are not path-sharing at first but are later on, only 730 interaction pairs are left. 
We characterize the closeness of these interactions in both distance and time in a density heatmap, see \Cref{fig:path_sharing_density}.
From the figure it becomes clear that the majority of interactions are close, i.e., the time difference is smaller than 6 seconds and the real-time closest distance is smaller than 20m. However, there is also a substantial part of path-sharing interactions, where there is a big time difference between the agents starting to occupy the same path or the distance between them is quite large.
Since we are interested in safety-critical interactions, the time difference between the agents should be relatively small.
Thus, we apply our third interaction criterion, i.e., $\Delta t_{\mathrm{PS}} \leq 6 \, s$, after which only 351 interaction pairs are left in the full train and validation split. That means only $2.1\%$ of the possible interactions are considered safety-critical. 

\begin{figure}
    \centering
    \includegraphics[width=0.45\textwidth]{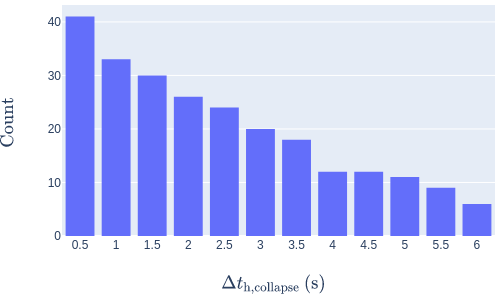}
    \caption{Histogram of data samples before the inevitable homotopy state. The samples are prediction frames of safety-critical interaction pairs in the nuScenes validation split.}
    \label{fig:histogram_dt_h_collapse}
\end{figure}

For testing the models on nuScenes, we evaluate them only on the validation split, which contains just 41 safety-critical interaction pairs.
After identifying which interactions to evaluate, we now determine when to evaluate them.
Employing our methodology for determining the inevitable homotopy state, we analyze the duration of the interaction interval $[t_\mathrm{h,start}, t_\mathrm{h,final}]$ before the homotopy class collapses. 
In \Cref{fig:histogram_dt_h_collapse}, we present a histogram showing the distribution of samples over their time to the inevitable homotopy state, $\Delta t_\mathrm{h,collapse}$.
Naturally, this histogram shows a decreasing trend, as the interval during which both agents are recorded in the dataset is relatively short for many interactions.
In total, we have just 41 usable interaction pairs, however, for the majority there are just a few samples available before the homotopy class becomes inevitable. There are just 6 pairs for which we can evaluate the predictions a full 6-second prediction horizon before $t_\mathrm{h,collapse}$.
Next, we will evaluate the models' mode prediction performance on these interaction pairs.


\subsection{Model intention prediction performance}
\label{subsec:interaction_metrics_results}

\begin{figure*}[hbt!]
    \centering
    \includegraphics[width=\textwidth]{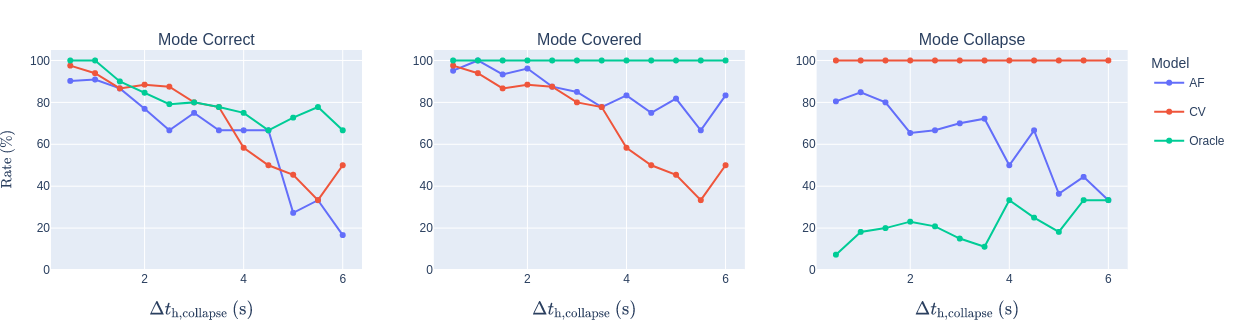}
    \caption{Relative mode prediction performance plotted against the time to inevitable homotopy state.
    From left to right, we consider the correct, covered and collapsed modes. We evaluate AF, the CV model, and the oracle on a prediction horizon of 6 seconds.}
    \label{fig:mode_metrics_6s_perDt_final}
\end{figure*}

\begin{table*}[hbt!]
    \centering
    \caption{Interaction mode prediction metrics for AF, CTT, the CV model and the oracle. 
    The rates are evaluated over all interaction-pair samples, whereas the time-based metrics and consistency are computed per interaction-pair sequence and later averaged. 
    We compare the mean time-to-correct/covered mode prediction, as well as the percentage of predictions that are correct from the beginning of the prediction interval ($@ T_\mathrm{pred}$) and the percentage of predictions where the $\Delta T = 0$, meaning the predictions are wrong even just before the inevitable homotopy collapse ($@0s$).
    The best metrics in each category are printed \textbf{bold} and the second-best \textit{italic}.
    }
    \label{tab:mode_metrics_combined}
    \begin{threeparttable}
    \begin{tabular}{@{\hskip 1mm}l@{\hskip 1mm}|c|c|c|c|ccc|c}
    \toprule
    \multicolumn{1}{c|}{\multirow{3}{*}[2pt]{Method}} & \multicolumn{1}{c|}{\multirow{3}{*}[2pt]{\shortstack{$T_\mathrm{pred}$ \\ $\mathrm{(s)}$}}} & \multirow{3}{*}{\shortstack{Mode correct \\ rate $\uparrow(\%)$}} & \multirow{3}{*}{\shortstack{Mode covered \\ rate  $\uparrow (\%)$}} & \multirow{3}{*}{\shortstack{Mode collapse \\ rate $\downarrow (\%)$}} & \multicolumn{3}{c|}{$\Delta T_{\mathrm{correct}}$ / $\Delta T_{\mathrm{covered}}$} & \multicolumn{1}{c}{\multirow{3}{*}[2pt]{\shortstack{Prediction \\ Consistency \\ $\uparrow (\%)$}}} \\ \cline{6-8}
    \rule{0pt}{3.5ex} & & & & & $\text{mean}\uparrow (s)$ & $ @ 0s \downarrow (\%) $ &  $ @ T_\mathrm{pred} \uparrow (\%) $ & \\ \midrule

    AF & \multirow{3}{*}{6} & 74.0 & \textit{89.3} & \textit{69.8} & 1.9 / 1.8 & 9.8 / 4.9 & 56.1 / \textit{80.5} & 92.7 \\
    CV model & & \textit{80.6} & 80.6 & 100.0 & \textit{2.3 / 2.3} & \textit{2.4 / 2.4} & \textbf{78.0} / 78.0 & \textbf{100.0} \\ 
    Oracle & & \textbf{86.0} & \textbf{100.0} & \textbf{18.6} & \textbf{2.4} / \textbf{-} & \textbf{0.0} / \textbf{0.0} & \textit{73.2} / \textbf{100.0} & \textit{97.6} \\ \midrule

    AF & \multirow{4}{*}{3} & 83.4 & \textit{92.9} & \textit{76.9} & \textit{1.0} / 0.8 & 12.2 / \textit{4.9} & \textit{70.7} / \textit{87.8} & \textit{95.1} \\
    CTT\tnote{*} & & 49.3 & 49.3 & 100.0 & 0.1 / 0.1 & 53.3 / 53.3 & 40.0 / 40.0 & \textbf{100.0} \\
    CV model & & \textbf{87.0} & 87.0 & 100.0 & 0.9 / \textit{0.9} & \textit{7.3} / 7.3 & \textbf{80.5} / 80.5 & \textbf{100.0} \\
    Oracle & & \textit{86.4} & \textbf{100.0} & \textbf{13.0} & \textbf{1.6} / \textbf{-} & \textbf{2.4} / \textbf{0.0} & \textit{70.7} / \textbf{100.0} & \textbf{100.0} \\
        \bottomrule
    \end{tabular}
      \begin{tablenotes}
      \item[*] Note that we were not able to reproduce the numbers reported in CTT's paper, and that some predictions are missing due to the issues discussed in \Cref{subsec:CTT}. 
      \end{tablenotes}
      \end{threeparttable}
\end{table*}

Predicting the driver's intentions 6 seconds before the interaction is far less important than predicting them 1 second before it happens. On the other hand, correctly predicting the intentions 1 second before the interaction happens, is also a lot easier, as the drivers in the scene have likely already implicitly communicated who takes priority and crosses first, resulting in increased margins and speed differences. 
To provide insights on the temporal evolution of a model's mode prediction performance, we analyze the mode correct, covered and collapse rates against the time to inevitable homotopy state $\Delta t_\mathrm{h,collapse}$, see \Cref{fig:mode_metrics_6s_perDt_final}.
Indeed, we see that, as the interaction comes closer (smaller $\Delta t_\mathrm{h,collapse}$), all models are able to more correctly predict the interaction class. 
That the alternative roll-outs are less likely to happen, is also reflected by the higher mode collapse rate of AF for samples closer to the inevitable homotopy state. Although, failing to cover a feasible mode when it is unlikely is not problematic. 
However, it is worth nothing that in some cases, the models are not even able to correctly predict the interaction class right before the homotopy class becomes inevitable, indicating that mode collapse also occurs in critical situations.

In \Cref{tab:mode_metrics_combined} the mode correctness, coverage and collapse rates of all models are summarized, as well as the time-based metrics and consistency. 
First, we will compare the intention prediction performance of AF, the oracle and the CV model on a prediction horizon of 6 seconds, and focus on the time-based metrics. 
AF correctly predicts the interaction mode at the beginning of the prediction horizon ($ @ T_\mathrm{pred}$) in 56\% of the cases, and if not, it takes up to 1.9 seconds on average to correctly predict the correct mode. Interestingly, the CV model outperforms the other models and in 78\% of the cases it can correctly predict the interaction mode from the beginning, by simply extrapolating the vehicle's current trajectory. This shows that in the majority of the cases, the interaction class is a natural evolution of the vehicle's current heading and speed. 

Naturally, in the covered category, the multimodal models perform better, as all predictions are considered. AF manages to directly cover the ground truth mode in 80\% of the cases, whereas the oracle model achieves a perfect score. 
The oracle model inherently tries to cover all feasible modes of the interacting agents, but since the models can only predict $K=5$ futures, it cannot completely mitigate mode collapse in scenes with many agents. The oracle scores a mode collapse rate of 19\%, versus the 70\% of AF. The CV model inherently suffers from 100\% mode collapse due to its unimodal predictions.  

Finally, we compare the models, including CTT, on a 3-second prediction horizon, with the results reported in the bottom half of \Cref{tab:mode_metrics_combined}. 
As explained earlier, CTT's predictions are unimodal and sometimes even missing, resulting in the model's disastrous performance. Only for 40\% of the interaction-pairs the mode is predicted correctly right away, and in 53\% of the cases, the mode is not predicted at all, resulting in a mean $\Delta T_{\mathrm{correct}}$ of 0.1 seconds. Because of the model's unimodal predictions, it inherently suffers from mode collapse for all scenarios. 
As we could not reproduce CTT's results, this is not representative of its real performance. However, by testing our methodology on multiple models, we show that it generalizes to other models.

Comparing the other models on the 6-second prediction horizon, we see that the results are slightly different because a shorter prediction horizon changes the number of samples and some predictions may fall in a different homotopy class for the shorter horizon.
However, the relative performance differences between the models remain unchanged. 
Although we cannot compare results from different prediction horizons directly, we demonstrated that our methodology is not limited to a single prediction horizon.

In terms of prediction consistency, all models score high: only in some cases the interaction mode changes inconsistently. The CV model and CTT score 100\%, which is more trivial, as they only output a single mode, so inconsistent mode predictions are less likely.

\subsection{Distance-based metrics results}
\label{subsec:distance_metrics}
In \Cref{tab:distance_metrics} we compare the models on the traditional distance-based metrics. 
We report the average and final displacement errors (ADE/FDE) for the most likely (ML) predictions, as well as the joint lower-bound metrics computed for $K=5$ modes.  
In contrast to our novel interaction metrics, these are computed over all scenes and time steps of the nuScenes validation split. 

\begin{table}[hbt!]
    \centering
    \caption{Distance-based metrics for AF, CTT, the CV model and the oracle, for 3-second and 6-second prediction horizons.
    The best metrics in each category are printed \textbf{bold} and the second-best \textit{italic}.
    }
    \label{tab:distance_metrics}
    \begin{threeparttable}
    \begin{tabular}{@{\hskip 1mm}l@{\hskip 1mm}|c|c|c|c|c}
    \toprule
    \multicolumn{1}{c|}{\multirow{3}{*}[2pt]{Method}} & \multicolumn{1}{c|}{\multirow{3}{*}[2pt]{\shortstack{$T_\mathrm{pred}$ \\ $\mathrm{(s)}$}}} & \multirow{2}{*}{\shortstack{ML \\ ADE \\$\downarrow$ (m)}} & \multirow{2}{*}{\shortstack{ML \\ FDE \\ $\downarrow$ (m)}} & \multirow{2}{*}{\shortstack{Joint \\ minADE \\ $\downarrow$ (m)}} & \multirow{2}{*}{\shortstack{Joint \\ minFDE \\ $\downarrow$ (m)}} \\ 
    \rule{0pt}{3.5ex} & & & & & \\ \midrule

    AF & \multirow{3}{*}{6} & 3.88 & \textit{9.10} & \textbf{2.86} & \textbf{6.48} \\
    CV model & & \textbf{3.64} & \textbf{9.04} & 3.64 & 9.04 \\ 
    Oracle & & \textit{3.84} & 9.12 & \textit{3.56} & \textit{8.41} \\ \midrule

    AF & \multirow{4}{*}{3} & 1.48 & 3.00 & \textbf{1.11} & \textbf{2.17} \\
    CTT\tnote{*} & & 5.93 & 10.59 & 5.93 & 10.59 \\
    CV model & & \textbf{1.22} & \textbf{2.68} & \textit{1.22} & 2.68 \\
    Oracle & & \textit{1.45} & \textit{2.85} & 1.36 & \textit{2.63} \\
    \bottomrule
    \end{tabular}
      \begin{tablenotes}
      \item[*] Note that we were not able to reproduce the numbers reported in CTT's paper, and that some predictions are missing due to the issues discussed in \Cref{subsec:CTT}. 
      \end{tablenotes}
      \end{threeparttable}
\end{table}

Comparing the ML ADE metrics to the ML interaction metrics, we see that the relative performance order remains similar, with the oracle and the CV model performing the best.
Interestingly, we see that on the joint metrics, AF performs best, which contradicts with our findings from the interaction metrics. 
This is partially caused by the fact that the oracle was designed specifically to cover modes of path-crossing vehicles, and not to get the lowest minimum distance errors.
More importantly, it also shows that in some cases, the joint lower-bound distance-based metrics fail to capture the model's ability to cover interaction modes among agent trajectories.

\section{CONCLUSION AND DISCUSSION}
\label{sec:conclusion_discussion}

We introduced a novel evaluation framework to benchmark a model's interaction prediction performance. 
Our framework simulates alternative interaction modes, and we use this to define a metric for mode collapse on the interaction level. 
We also use metrics for mode correctness and coverage, and propose time-based variants, that provide insight into the temporal evolution of mode predictions.  
Uniquely, our method does not evaluate all scenes and frames of a dataset, but only the relevant frames for closely interacting agent-pairs. 
This reduces the dataset dependency and makes our metrics more insightful and interpretable. 
We tested four models on the nuScenes dataset and showed that mode collapse happens. 
Interestingly, a simple constant velocity model outperformed the other models in correctly predicting the interaction mode, showing that in many cases the interaction mode is dictated by the vehicles' current heading and speed.
While AgentFormer (AF) manages to produce diverse predictions for each agent, it did not cover all feasible interaction modes between the interacting agents, averaging a mode collapse rate of 70\% for the safety-critical interaction pairs. 
The oracle model, designed to cover all feasible interaction modes, had a mode collapse rate of 20\%. Thus, completely alleviating mode collapse (i.e., covering all feasible interaction modes) is not possible with a finite number of \(K=5\) joint predictions due to the exponentially growing cardinality of the mode space. 
Although the oracle was superior in covering the interaction modes, it was outperformed by AF on the widely-used joint distance-based metrics, indicating that these metrics do not necessarily capture the model's performance in predicting interaction modes.
Finally, we analyzed the temporal evolution of the predictions, and found that both the mode correct and collapse rate increase as the inevitable homotopy state comes closer. 
In the majority of the scenarios, these collapsed interaction modes do not seem problematic, as they are not likely to happen. However, in a few cases, the models are not able to correctly predict the real interaction class just before it interaction settles. 
These incorrectly predicted driver intentions could pose safety concerns for autonomous driving. 

While we show that mode collapse occurs, our metrics do not evaluate the severity of consequences, nor the likelihood, of the collapsed modes. 
In our framework, we simulate feasible futures for interacting agents at every time step, but the model inputs remain the ground truth history of the agents as we replay the scene.
Assessing the safety implications of collapsed modes requires a closed-loop simulation setup, in which the predictions are used in a downstream planner.
Estimating the likelihood of a collapsed mode could involve comparing the scenario to a distribution learned from traffic data. However, rare but feasible scenarios might be underrepresented and deemed unlikely. 
Alternatively, planning-like costs could be used to evaluate the safety, comfort, and utility of future roll-outs as a measure of probability. 
Extending our framework to assess the associated risks of false predictions presents an exciting opportunity for future research.




In our framework, we only perform roll-outs and collision checks for pairs of interacting agents. However, in reality, the scenes can be more complex, with multiple agents interacting and influencing each other. 
While this is a conceptual limitation of our method, the feasibility of our simulations remains valid, as the feasibility is primarily determined by the yielding vehicle's ability to brake before entering the common path, which is not affected by other vehicles.  

By applying our methodology to identify safety-critical interactions, we make the evaluation less dependent on the dataset while focusing on the most crucial aspect of driving: the interactions. 
In the nuScenes dataset, we found that only 2\% of the theoretical interactions are considered safety-critical according to our criteria.
This finding highlights the need for more interactive datasets and the importance of metrics that are less constrained on the scenarios in a dataset. 
However, it also reveals a limitation of our approach: we evaluate only real interactions, not hypothetical ones. We opted for this simplistic approach to ensure that the interactions we assess are realistic. Furthermore, simulating all hypothetical interactions would be extremely complex and computationally demanding.

Finally, we analyzed the temporal evolution of the interactions between the critical agent-pairs. 
For the majority of the pairs, however, there were only a few samples available prior to the interaction, limiting the interpretability of our time-based metrics.
This limitation arises because nuScenes is recorded from an on-road viewpoint, constraining the annotations to the range of the ego vehicle. To address this issue, future research could apply our methodology to traffic datasets recorded from a top-down perspective, such as those captured by drones monitoring traffic at intersections \cite{krajewski_round_2020,bock_ind_2020, zhan_interaction_2019}.




Our novel interaction metrics provide new ways to measure the intention prediction of models in safety-critical interactions.
These metrics only take into account the relevant interactions, thereby reducing the dependency on datasets and improving interpretability.
Furthermore, our time-based metrics shed light on the temporal evolution of predictions, an aspect that was previously neglected in VTP evaluation. 
Our new evaluation methodology thus offers new insights and perspectives, helping the holistic evaluation and interpretation of a model's performance. 
Finally, our evaluation methodology can aid the development of VTP models towards more accurate and consistent interaction predictions. 
Future work should focus on alleviating the aforementioned weaknesses and further generalizing our framework to other datasets and models to establish a benchmark for prediction models.


\section*{ACKNOWLEDGMENT}
The authors would like to thank Dr. C. Pek for his helpful comments and feedback, and I. van Vorst for proofreading the paper. 


%

\bibliographystyle{IEEEtran}
\bibliography{references}

\begin{thebibliography}{10}
\providecommand{\url}[1]{#1}
\csname url@samestyle\endcsname
\providecommand{\newblock}{\relax}
\providecommand{\bibinfo}[2]{#2}
\providecommand{\BIBentrySTDinterwordspacing}{\spaceskip=0pt\relax}
\providecommand{\BIBentryALTinterwordstretchfactor}{4}
\providecommand{\BIBentryALTinterwordspacing}{\spaceskip=\fontdimen2\font plus
\BIBentryALTinterwordstretchfactor\fontdimen3\font minus \fontdimen4\font\relax}
\providecommand{\BIBforeignlanguage}[2]{{%
\expandafter\ifx\csname l@#1\endcsname\relax
\typeout{** WARNING: IEEEtran.bst: No hyphenation pattern has been}%
\typeout{** loaded for the language `#1'. Using the pattern for}%
\typeout{** the default language instead.}%
\else
\language=\csname l@#1\endcsname
\fi
#2}}
\providecommand{\BIBdecl}{\relax}
\BIBdecl

\bibitem{othman_exploring_2022}
\BIBentryALTinterwordspacing
K.~Othman, ``Exploring the implications of autonomous vehicles: a comprehensive review,'' \emph{Innovative Infrastructure Solutions}, vol.~7, no.~2, p. 165, 2022. [Online]. Available: \url{https://www.ncbi.nlm.nih.gov/pmc/articles/PMC8885781/}
\BIBentrySTDinterwordspacing

\bibitem{hagedorn_rethinking_2023}
S.~Hagedorn, M.~Hallgarten, M.~Stoll, and A.~P. Condurache, ``The integration of prediction and planning in deep learning automated driving systems: A review,'' \emph{IEEE Transactions on Intelligent Vehicles}, pp. 1--17, 2024.

\bibitem{gupta_social_2018}
A.~Gupta, J.~Johnson, L.~Fei-Fei, S.~Savarese, and A.~Alahi, ``\BIBforeignlanguage{en}{Social {GAN}: {Socially} {Acceptable} {Trajectories} with {Generative} {Adversarial} {Networks}},'' in \emph{\BIBforeignlanguage{en}{2018 {IEEE}/{CVF} {Conference} on {Computer} {Vision} and {Pattern} {Recognition}}}.\hskip 1em plus 0.5em minus 0.4em\relax Salt Lake City, UT: IEEE, Jun. 2018, pp. 2255--2264.

\bibitem{amirian_social_2019}
J.~Amirian, J.-B. Hayet, and J.~Pettre, ``\BIBforeignlanguage{en}{Social {Ways}: {Learning} {Multi}-{Modal} {Distributions} of {Pedestrian} {Trajectories} {With} {GANs}},'' in \emph{\BIBforeignlanguage{en}{2019 {IEEE}/{CVF} {Conference} on {Computer} {Vision} and {Pattern} {Recognition} {Workshops} ({CVPRW})}}.\hskip 1em plus 0.5em minus 0.4em\relax Long Beach, CA, USA: IEEE, Jun. 2019, pp. 2964--2972.

\bibitem{zhao_tnt_2021}
H.~Zhao, J.~Gao, T.~Lan, C.~Sun, B.~Sapp, B.~Varadarajan, Y.~Shen, Y.~Shen, Y.~Chai, C.~Schmid, C.~Li, and D.~Anguelov, ``\BIBforeignlanguage{en}{{TNT}: {Target}-driven {Trajectory} {Prediction}},'' in \emph{\BIBforeignlanguage{en}{Proceedings of the 2020 {Conference} on {Robot} {Learning}}}.\hskip 1em plus 0.5em minus 0.4em\relax PMLR, Oct. 2021, pp. 895--904, iSSN: 2640-3498.

\bibitem{gu_densetnt_2021}
J.~Gu, C.~Sun, and H.~Zhao, ``\BIBforeignlanguage{en}{{DenseTNT}: {End}-to-end {Trajectory} {Prediction} from {Dense} {Goal} {Sets}},'' in \emph{\BIBforeignlanguage{en}{2021 {IEEE}/{CVF} {International} {Conference} on {Computer} {Vision} ({ICCV})}}.\hskip 1em plus 0.5em minus 0.4em\relax Montreal, QC, Canada: IEEE, Oct. 2021, pp. 15\,283--15\,292.

\bibitem{chen_categorical_2023}
Y.~Chen, S.~Tonkens, and M.~Pavone, ``Categorical {Traffic} {Transformer}: {Interpretable} and {Diverse} {Behavior} {Prediction} with {Tokenized} {Latent},'' Nov. 2023, arXiv:2311.18307 [cs].

\bibitem{chen_scept_2022}
Y.~Chen, B.~Ivanovic, and M.~Pavone, ``\BIBforeignlanguage{en}{{ScePT}: {Scene}-consistent, {Policy}-based {Trajectory} {Predictions} for {Planning}},'' in \emph{\BIBforeignlanguage{en}{2022 {IEEE}/{CVF} {Conference} on {Computer} {Vision} and {Pattern} {Recognition} ({CVPR})}}.\hskip 1em plus 0.5em minus 0.4em\relax New Orleans, LA, USA: IEEE, Jun. 2022, pp. 17\,082--17\,091.

\bibitem{yuan_dlow_2020}
Y.~Yuan and K.~Kitani, ``\BIBforeignlanguage{en}{{DLow}: {Diversifying} {Latent} {Flows} for {Diverse} {Human} {Motion} {Prediction}},'' in \emph{\BIBforeignlanguage{en}{Computer {Vision} – {ECCV} 2020}}, A.~Vedaldi, H.~Bischof, T.~Brox, and J.-M. Frahm, Eds.\hskip 1em plus 0.5em minus 0.4em\relax Cham: Springer International Publishing, 2020, pp. 346--364.

\bibitem{deo_trajectory_2021}
\BIBentryALTinterwordspacing
N.~Deo and M.~M. Trivedi, ``Trajectory {Forecasts} in {Unknown} {Environments} {Conditioned} on {Grid}-{Based} {Plans},'' Apr. 2021, arXiv:2001.00735 [cs]. [Online]. Available: \url{http://arxiv.org/abs/2001.00735}
\BIBentrySTDinterwordspacing

\bibitem{ajanovic_search-based_2018}
Z.~Ajanović, B.~Lacevic, B.~Shyrokau, M.~Stolz, and M.~Horn, ``Search-{Based} {Optimal} {Motion} {Planning} for {Automated} {Driving},'' Oct. 2018.

\bibitem{berkemeyer_feasible_2021}
H.~Berkemeyer, R.~Franceschini, T.~Tran, L.~Che, and G.~Pipa, ``Feasible and {Adaptive} {Multimodal} {Trajectory} {Prediction} with {Semantic} {Maneuver} {Fusion},'' in \emph{2021 {IEEE} {International} {Conference} on {Robotics} and {Automation} ({ICRA})}, May 2021, pp. 8530--8536, iSSN: 2577-087X.

\bibitem{kumar_interaction-based_2021}
S.~Kumar, Y.~Gu, J.~Hoang, G.~C. Haynes, and M.~Marchetti-Bowick, ``Interaction-{Based} {Trajectory} {Prediction} {Over} a {Hybrid} {Traffic} {Graph},'' in \emph{2021 {IEEE}/{RSJ} {International} {Conference} on {Intelligent} {Robots} and {Systems} ({IROS})}, Sep. 2021, pp. 5530--5535, iSSN: 2153-0866.

\bibitem{chen_interactive_2024}
Y.~Chen, S.~Veer, P.~Karkus, and M.~Pavone, ``Interactive {Joint} {Planning} for {Autonomous} {Vehicles},'' \emph{IEEE Robotics and Automation Letters}, vol.~9, no.~2, pp. 987--994, Feb. 2024, conference Name: IEEE Robotics and Automation Letters.

\bibitem{chen_tree-structured_2023}
Y.~Chen, P.~Karkus, B.~Ivanovic, X.~Weng, and M.~Pavone, ``Tree-structured {Policy} {Planning} with {Learned} {Behavior} {Models},'' in \emph{2023 {IEEE} {International} {Conference} on {Robotics} and {Automation} ({ICRA})}, May 2023, pp. 7902--7908.

\bibitem{luo_jfp_2023}
\BIBentryALTinterwordspacing
W.~Luo, C.~Park, A.~Cornman, B.~Sapp, and D.~Anguelov, ``\BIBforeignlanguage{en}{{JFP}: {Joint} {Future} {Prediction} with {Interactive} {Multi}-{Agent} {Modeling} for {Autonomous} {Driving}},'' in \emph{\BIBforeignlanguage{en}{Proceedings of {The} 6th {Conference} on {Robot} {Learning}}}.\hskip 1em plus 0.5em minus 0.4em\relax PMLR, Mar. 2023, pp. 1457--1467, iSSN: 2640-3498. [Online]. Available: \url{https://proceedings.mlr.press/v205/luo23a.html}
\BIBentrySTDinterwordspacing

\bibitem{yuan_agentformer_2021}
Y.~Yuan, X.~Weng, Y.~Ou, and K.~Kitani, ``\BIBforeignlanguage{en}{{AgentFormer}: {Agent}-{Aware} {Transformers} for {Socio}-{Temporal} {Multi}-{Agent} {Forecasting}},'' in \emph{\BIBforeignlanguage{en}{2021 {IEEE}/{CVF} {International} {Conference} on {Computer} {Vision} ({ICCV})}}.\hskip 1em plus 0.5em minus 0.4em\relax Montreal, QC, Canada: IEEE, Oct. 2021, pp. 9793--9803.

\bibitem{ettinger_large_2021}
S.~Ettinger, S.~Cheng, B.~Caine, C.~Liu, H.~Zhao, S.~Pradhan, Y.~Chai, B.~Sapp, C.~Qi, Y.~Zhou, Z.~Yang, A.~Chouard, P.~Sun, J.~Ngiam, V.~Vasudevan, A.~McCauley, J.~Shlens, and D.~Anguelov, ``Large {Scale} {Interactive} {Motion} {Forecasting} for {Autonomous} {Driving} : {The} {Waymo} {Open} {Motion} {Dataset},'' in \emph{2021 {IEEE}/{CVF} {International} {Conference} on {Computer} {Vision} ({ICCV})}, Oct. 2021, pp. 9690--9699, iSSN: 2380-7504.

\bibitem{markkula_defining_2020}
G.~Markkula, R.~Madigan, D.~Nathanael, E.~Portouli, Y.~M. Lee, A.~Dietrich, J.~Billington, A.~Schieben, and N.~Merat, ``Defining interactions: a conceptual framework for understanding interactive behaviour in human and automated road traffic,'' \emph{Theoretical Issues in Ergonomics Science}, vol.~21, no.~6, pp. 728--752, Nov. 2020, publisher: Taylor \& Francis.

\bibitem{bhattacharya_topological_2012}
S.~Bhattacharya, M.~Likhachev, and V.~Kumar, ``\BIBforeignlanguage{English}{Topological constraints in search-based robot path planning},'' \emph{\BIBforeignlanguage{English}{Autonomous Robots}}, vol.~3, no.~33, pp. 273--290, 2012.

\bibitem{ziegler_fast_2010}
J.~Ziegler and C.~Stiller, ``\BIBforeignlanguage{en}{Fast collision checking for intelligent vehicle motion planning},'' in \emph{\BIBforeignlanguage{en}{2010 {IEEE} {Intelligent} {Vehicles} {Symposium}}}.\hskip 1em plus 0.5em minus 0.4em\relax La Jolla, CA, USA: IEEE, Jun. 2010, pp. 518--522.

\bibitem{caesar_nuscenes_2020}
H.~Caesar, V.~Bankiti, A.~H. Lang, S.~Vora, V.~E. Liong, Q.~Xu, A.~Krishnan, Y.~Pan, G.~Baldan, and O.~Beijbom, ``{nuScenes}: {A} {Multimodal} {Dataset} for {Autonomous} {Driving},'' in \emph{2020 {IEEE}/{CVF} {Conference} on {Computer} {Vision} and {Pattern} {Recognition} ({CVPR})}, Jun. 2020, pp. 11\,618--11\,628, iSSN: 2575-7075.

\bibitem{yuan_khrylxagentformer_2024}
\BIBentryALTinterwordspacing
Y.~Yuan, ``Khrylx/{AgentFormer},'' May 2024, original-date: 2021-03-24T16:40:46Z. [Online]. Available: \url{https://github.com/Khrylx/AgentFormer}
\BIBentrySTDinterwordspacing

\bibitem{noauthor_nvlabsdiffstack_nodate}
\BIBentryALTinterwordspacing
``{NVlabs}/diffstack at {CTT} release.'' [Online]. Available: \url{https://github.com/NVlabs/diffstack/tree/CTT_release}
\BIBentrySTDinterwordspacing

\bibitem{karle_scenario_2022}
P.~Karle, M.~Geisslinger, J.~Betz, and M.~Lienkamp, ``\BIBforeignlanguage{English}{Scenario {Understanding} and {Motion} {Prediction} for {Autonomous} {Vehicles} - {Review} and {Comparison}},'' \emph{\BIBforeignlanguage{English}{IEEE Transactions on Intelligent Transportation Systems}}, vol.~23, no.~10, pp. 16\,962--16\,982, 2022.

\bibitem{de_winkel_standards_2023}
K.~N. de~Winkel, T.~Irmak, R.~Happee, and B.~Shyrokau, ``Standards for passenger comfort in automated vehicles: {Acceleration} and jerk,'' \emph{Applied Ergonomics}, vol. 106, p. 103881, Jan. 2023.

\bibitem{krajewski_round_2020}
\BIBentryALTinterwordspacing
R.~Krajewski, T.~Moers, J.~Bock, L.~Vater, and L.~Eckstein, ``The {rounD} {Dataset}: {A} {Drone} {Dataset} of {Road} {User} {Trajectories} at {Roundabouts} in {Germany},'' in \emph{2020 {IEEE} 23rd {International} {Conference} on {Intelligent} {Transportation} {Systems} ({ITSC})}, Sep. 2020, pp. 1--6. [Online]. Available: \url{https://ieeexplore.ieee.org/document/9294728}
\BIBentrySTDinterwordspacing

\bibitem{bock_ind_2020}
J.~Bock, R.~Krajewski, T.~Moers, S.~Runde, L.~Vater, and L.~Eckstein, ``The {inD} {Dataset}: {A} {Drone} {Dataset} of {Naturalistic} {Road} {User} {Trajectories} at {German} {Intersections},'' in \emph{2020 {IEEE} {Intelligent} {Vehicles} {Symposium} ({IV})}, Oct. 2020, pp. 1929--1934, iSSN: 2642-7214.

\bibitem{zhan_interaction_2019}
W.~Zhan, L.~Sun, D.~Wang, H.~Shi, A.~Clausse, M.~Naumann, J.~Kummerle, H.~Konigshof, C.~Stiller, A.~de~La~Fortelle, and M.~Tomizuka, ``{INTERACTION} {Dataset}: {An} {INTERnational}, {Adversarial} and {Cooperative} {moTION} {Dataset} in {Interactive} {Driving} {Scenarios} with {Semantic} {Maps},'' Sep. 2019, arXiv:1910.03088 [cs, eess].

\end{thebibliography}






\begin{IEEEbiography}[{\includegraphics[trim=15cm 10cm 18cm 15cm, clip=true,width=1in,height=1.25in,clip,keepaspectratio]{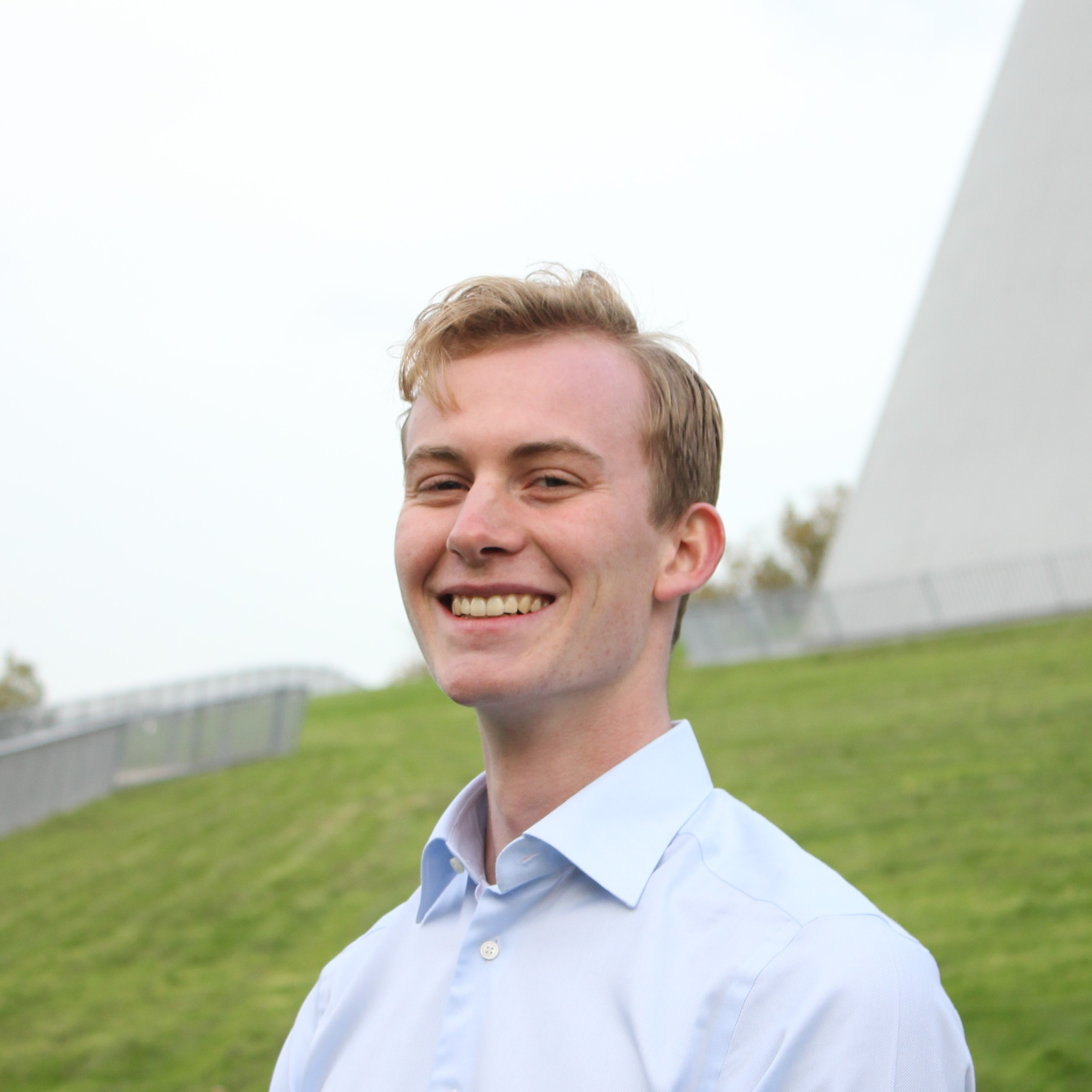}}]{Maarten Hugenholtz} received his BSc degree in Mechanical Engineering from Delft University of Technology in 2020. After spending two years in Formula Student and Formula E, he graduated in 2024 from the MSc Robotics, focusing on the evaluation of mode collapse in trajectory prediction models. 
\end{IEEEbiography}

\begin{IEEEbiography}[{\includegraphics[width=1in,height=1.25in,clip,keepaspectratio]{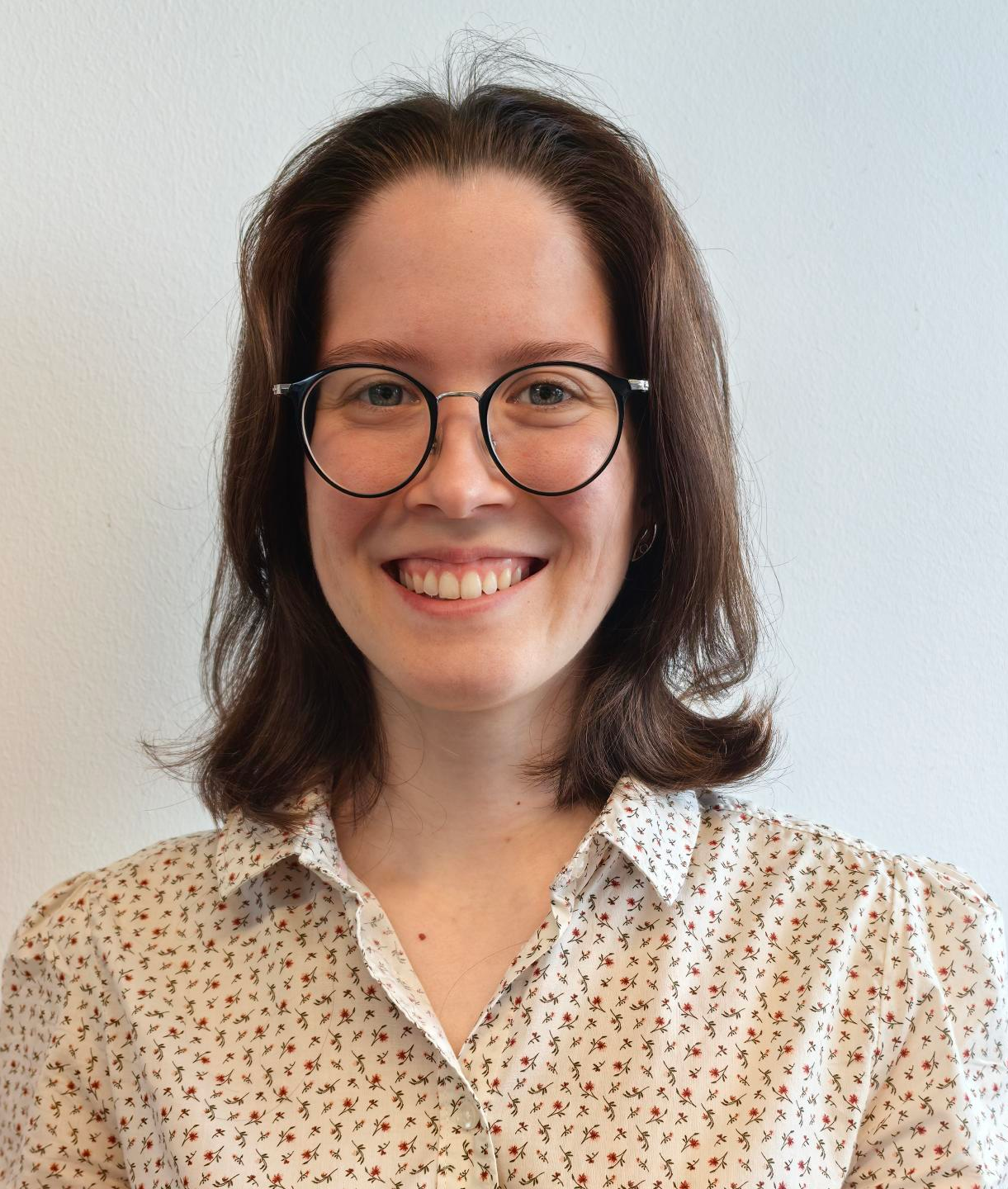}}]{Anna M\'esz\'aros} is a PhD student at the Cognitive Robotics Department at TU Delft, Netherlands since 2022. She received a BSc in Robotics and Automation at the University of L{\"u}beck, Germany in 2019 and an MSc in Mechanical Engineering with a specialisation in Robotics at TU Delft in 2021. Her research focuses on probabilistic prediction of traffic participants in urban environments.
\end{IEEEbiography}

\begin{IEEEbiography}[{\includegraphics[width=1in,height=1.25in,clip,keepaspectratio]{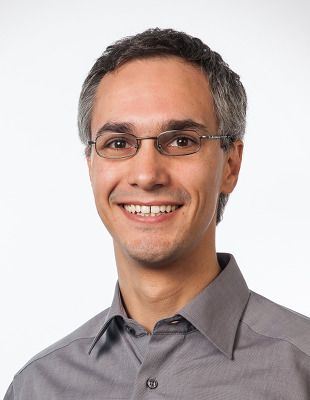}}]{Jens Kober} Jens Kober received the Ph.D. degree in engineering from TU Darmstadt, Darmstadt, Germany, in 2012. He is currently an Associate Professor with TU Delft, Delft, The Netherlands. He was a Postdoctoral Scholar jointly with CoR-Lab, Bielefeld University, Bielefeld, Germany, and with Honda Research Institute Europe, Offenbach, Germany.
Dr. Kober was a recipient of the annually awarded Georges Giralt PhD Award for the best PhD thesis in robotics in Europe, the 2018 IEEE RAS Early Academic Career Award, the 2022 RSS Early Career Award, and was a recipient of an ERC Starting grant.
\end{IEEEbiography}

\begin{IEEEbiography}[{\includegraphics[width=1in,height=1.25in,clip,keepaspectratio]{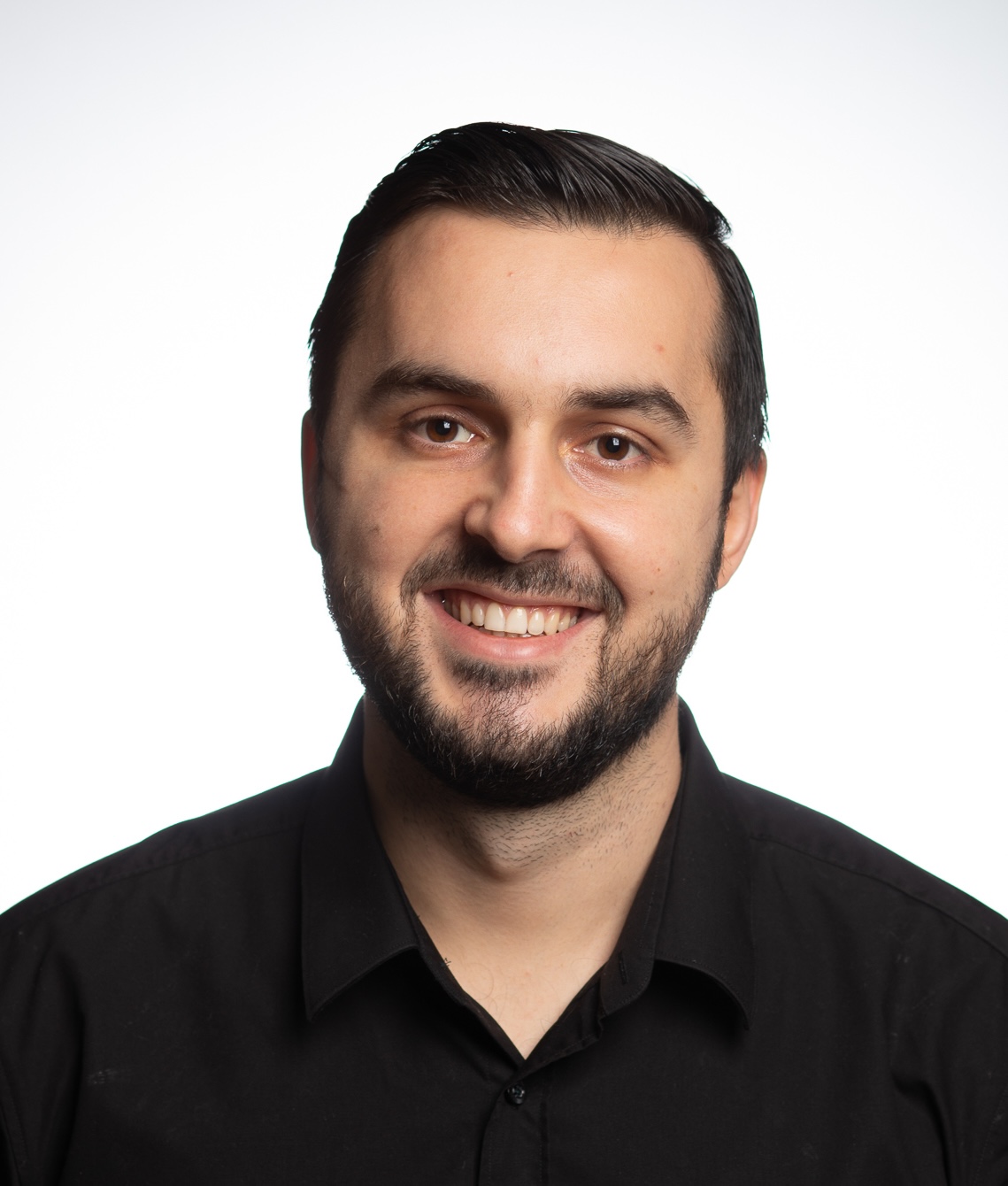}}]{Zlatan Ajanovi{\'c}} is a postdoctoral Researcher at RWTH Aachen University, with Prof. Hector Geffner. Before, he was a postdoctoral researcher at TU Delft with Prof. Jens Kober. He earned PhD degree from the Graz University of Technology. His main research focus is on Decision-making and Control for Autonomous Robots based on Planning, Learning, and Control Theory. He received the IFAC Young Author Award, Hans List Award, and DAAD AInet PostDoc Fellowship.
\end{IEEEbiography}



\vfill

\end{document}